\documentclass[journal,twoside,web]{ieeecolor}
\usepackage{generic}
\usepackage{cite}
\usepackage{amsmath,amssymb,amsfonts}
\usepackage{algorithmic}
\usepackage{graphicx}
\usepackage{algorithm,algorithmic}
\usepackage{hyperref}
\hypersetup{hidelinks=true}
\usepackage{textcomp}

\usepackage{colortbl}
\usepackage{xcolor}
\usepackage{subcaption}
\usepackage{booktabs}
\usepackage{multirow}
\usepackage{pifont}
\usepackage{wrapfig}
\usepackage[misc]{ifsym}

\usepackage{xspace}
\newcommand{\editorial}[2]{{\color{#1}{#2}}\xspace}
\newcommand{\jbhi}[1]{\editorial{black}{#1}}
\usepackage{orcidlink}

\def\BibTeX{{\rm B\kern-.05em{\sc i\kern-.025em b}\kern-.08em
    T\kern-.1667em\lower.7ex\hbox{E}\kern-.125emX}}
\begin{document}
\title{Gaussian Primitives for Deformable Image Registration}
\author{
Jihe Li*\orcidlink{0009-0006-4575-0026}, Xiang Liu*\orcidlink{0009-0006-8550-3767}, Fabian Zhang\orcidlink{0009-0003-6507-8204}, Xia Li\textsuperscript{\Letter}\orcidlink{0000-0003-2284-1700}, Xixin Cao\orcidlink{0009-0002-1090-2844},
Ye Zhang\orcidlink{0000-0003-1608-4467}, and Joachim Buhmann\orcidlink{0000-0002-6613-7101} 
\thanks{Manuscript received 14 October 2024. This project is supported by the interdisciplinary doctoral grant (iDoc 2021-360) from the Personalized Health and Related Technologies (PHRT) of the ETH domain, Switzerland. \emph{*Joint first authors. \textsuperscript{\Letter}Corresponding author.}}
\thanks{Jihe Li and Xixin Cao are with School of Software and Microelectronics, Peking University, Beijing 100871, China (e-mail: lijh@stu.pku.edu.cn, cxx@ss.pku.edu.cn)}
\thanks{Xiang Liu is with School of Computing, National University of Singapore, Singapore 119077 (e-mail: liuxiang@comp.nus.edu.sg)}
\thanks{Fabian Zhang, Xia Li, and Joachim Buhmann are with Department of Computer Science, ETH Zurich, Zurich 8092, Switzerland (e-mail: fabzhang@ethz.ch, ethlixia@gmail.com, jbuhmann@inf.ethz.ch)}
\thanks{Xia Li and Ye Zhang are with Center for Proton Therapy, Paul Scherrer Institut, Villigen 5232, Switzerland (e-mail: ethlixia@gmail.com, ye.zhang@psi.ch)}
}

\maketitle

\begin{abstract}
Deformable Image Registration (DIR) is essential for aligning medical images that exhibit anatomical variations, facilitating applications such as disease tracking and radiotherapy planning. While classical iterative methods and deep learning approaches have achieved success in DIR, they are often hindered by computational inefficiency or poor generalization. In this paper, we introduce GaussianDIR, a novel, case-specific optimization DIR method inspired by 3D Gaussian splatting.
In general, GaussianDIR represents image deformations using a sparse set of mobile and flexible Gaussian primitives, each defined by a center position, covariance, and local rigid transformation.
This compact and explicit representation reduces noise and computational overhead while improving interpretability.
Furthermore, the movement of individual voxel is derived via blending the local rigid transformation of the neighboring Gaussian primitives. By this, GaussianDIR captures both global smoothness and local rigidity as well as reduces the computational burden. 
To address varying levels of deformation complexity, GaussianDIR also integrates an adaptive density control mechanism that dynamically adjusts the density of Gaussian primitives.
Additionally, we employ multi-scale Gaussian primitives to capture both coarse and fine deformations, reducing optimization to local minima.
Experimental results on brain MRI, lung CT, and cardiac MRI datasets demonstrate that GaussianDIR outperforms existing DIR methods in both accuracy and efficiency, highlighting its potential for clinical applications.
Finally, as a training-free approach, it challenges the stereotype that iterative methods are inherently slow and transcend the limitations of poor generalization.

\end{abstract}

\section{Introduction}\label{introduction}
\PARstart{D}{eformable} Image Registration (DIR) is fundamental in medical imaging, allowing for the alignment of \jbhi{medical} images with significant anatomical variations. Given a pair of images (fixed and moving), DIR aims to compute a displacement vector field (DVF) that warps the moving image to align with the fixed image. Unlike rigid or affine registration, DIR can handle complex, non-linear deformations, making it indispensable for applications such as contour propagation~\cite{hautvast2006automatic}, dose accumulation~\cite{ibragimov2019neural}, and radiotherapy treatment planning~\cite{ren2013treatment, commandeur2016mri, shao2020prediction}. By accurately aligning corresponding anatomical structures, DIR enhances diagnostic precision, facilitates disease progression tracking, and improves the accuracy of interventions. Its ability to account for patient-specific variability is essential for personalized medicine and adaptive therapeutic \jbhi{diagnosis and treatment recommendations}. 

Various approaches have been proposed for DIR, which can be broadly categorized into classical iterative optimization-based, deep learning (DL)-based, and implicit neural representations (INR)-based methods. 
Classical iterative methods~\cite{thirion1998image, cao2005large, vishnevskiy2016isotropic, siebert2021fast, jena2024fireants} formulate DIR as a variational optimization problem that relies heavily on mathematical priors. These methods are robust and generalize well across various modalities, while they often suffer from high computational costs. For instance, LDDMM~\cite{lddmm} requires solving complex partial differential equations, leading to significant computational overhead, especially when processing images with large size~\cite{castillo2009framework}. ConvexAdam~\cite{siebert2021fast} attempt to accelerate the process by optimizing the DVF at a coarse resolution before upsampling it, but this compromise often leads to reduced accuracy.

The rise of DL has introduced AI data-driven approaches that leverage large datasets to train convolutional neural networks (CNNs) or Transformer to predict dense DVF from pairs of fixed and moving images~\cite{balakrishnan2019voxelmorph, hoopes2021hypermorph, hoffmann2021synthmorph, chen2022transmorph, chen2023transmatch}. 
One advantage of DL-based methods is their ability to automatically extract features, coping with the limitation of the fidelity of image intensities~\cite{jena2024deep}. Another benefit is to incorporate weak supervision from anatomical landmarks or segmentation maps ~\cite{chen2022transmorph, jena2024deep} during training. 
However, DL-based methods are limited by their dependency on large datasets, which are often difficult and expensive to obtain in medical settings. Moreover, they tend to struggle with generalization when applied to data distributions different from the training set~\cite{jena2024deep}.

INR-based methods~\cite{wolterink2022implicit, van2023robust,li2024continuous} use multi-layer perceptrons (MLPs) to represent the coordinates mapping function between fixed and moving images. 
One of their key designs is to perform mini-batch gradient descent to optimize the network, enabling relatively fast processing despite facing high-resolution DVF.
However, INR-based methods often suffer from challenge in capturing sharp and complex deformations and lack interpretability due to their implicit nature, which still lack real-time speed. Recently, Gaussian Splatting (GS)~\cite{kerbl20233d} has emerged as a more efficient and interpretable approach for 3D reconstruction, using mobile and elastic 3D Gaussian together with spherical harmonics (SH) to model appearance of local regions. For improving rendering efficiency, it utilizes truncated Gaussian to perform $\alpha$-blending. 
Inspired by 3D GS, we innovatively explore the application of Gaussian primitives in DIR. Our aim is to combine registration efficiency and generalization, which existing methods struggle to achieve, while enhancing interpretability compared to INR-based and DL-based methods.

This study introduces GaussianDIR, a novel case-specific optimization DIR approach beyond traditional grid-based~\cite{modat2010fast, vishnevskiy2016isotropic} or voxel-based representations~\cite{thirion1998image, vercauteren2009diffeomorphic, shi2012registration}.
GaussianDIR employs a set of mobile and elastic 3D Gaussian primitives defined by their center positions and covariance. Each Gaussian is assigned a local rigid transformation to model motion instead of SH coefficient for appearance, providing a compact and flexible representation for DVF. 
A local region of anatomical tissue tends to require similar deformation, thus we leverage a Gaussian primitive to represent the deformation of the local area.
This formulation allows GaussianDIR to represent deformations more efficiently and with fewer parameters than voxel-based methods, while also avoiding overfitting to noise.
Besides, this explicit representation utilizing Gaussian primitives effectively improves the interpretability, overcoming the limitations inherent in INR.

GaussianDIR determines the motion of each voxel by blending the local rigid transformations of its $K$ nearest Gaussian primitives, identified using the $K$-nearest neighbors (KNN) algorithm. 
This strategy mimics the truncated Gaussian approach in 3D GS~\cite{kerbl20233d}, allowing us to not only capture both global smoothness and local rigidity of anatomical structures but also reduce the computational burden in transformation blending. Additionally, we integrate an adaptive mechanism that dynamically adjusts the density of Gaussian primitives to handle varying levels of deformation complexity.
Compared to traditional B-spline-based methods that use fixed control points on a grid, GaussianDIR’s dynamic and adaptive nature provides more flexibility. Furthermore, we employ multi-scale Gaussian primitives to address both coarse and fine deformations, mitigating the risk of converging to suboptimal local minima during optimization.

We conduct extensive experiments and rigorous evaluation on brain MRI (OASIS~\cite{marcus2007open}), lung CT (DIRLab~\cite{castillo2009framework}), and cardiac MRI (ACDC~\cite{bernard2018deep}) benchmarks. The results demonstrate that GaussianDIR not only outperforms existing methods both qualitatively and quantitatively but also achieves a significant speedup when processing large medical images \jbhi{automatically}, making it highly practical for clinical usage.

\begin{figure*}[htbp]
    \centering
    \includegraphics[width=\textwidth]{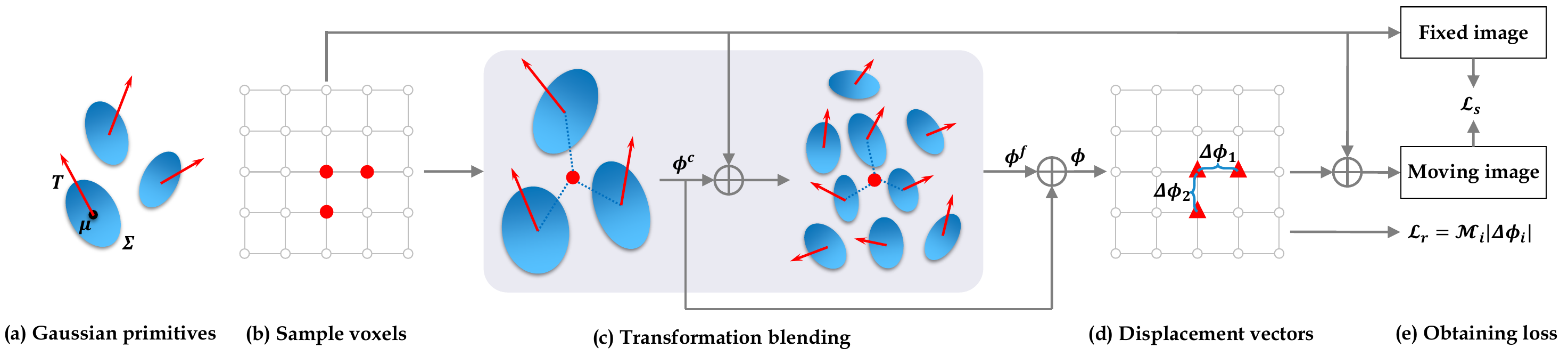}
    \caption{Pipeline of GaussianDIR: (a) Blue ellipses represent 3D Gaussians parameterized by center position $\boldsymbol{\mu}$ and covariance matrix $\boldsymbol{\Sigma}$. Red arrows indicate the local transformations $\boldsymbol{T}$. (b) Voxel groups are sampled from the volume to enable mini-batch optimization and total variation regularization. (c) Local transformations of neighboring Gaussian primitives are blended, with neighbors identified using the KNN algorithm at two different scales. (d) Deformation vectors $\boldsymbol{\phi}$ are obtained after transformation blending. (e) The similarity loss $\mathcal{L}_s$ and regularization loss $\mathcal{L}_r$ are then calculated. $\mathcal{M}$ represents the mean operation.}
    \label{fig:framework}
\end{figure*}

\section{Related Works}\label{relatedwork}
\subsection{Classical Methods} 

Classical image registration algorithms are well-established and grounded in robust mathematical theory. The Demons algorithm~\cite{thirion1998image} utilizes the optical flow equation to iteratively align images by modeling voxel intensity differences as forces, followed by the emergence of its diffeomorphic variants~\cite{vercauteren2009diffeomorphic}. The Symmetric Normalization (SyN) algorithm~\cite{avants2008symmetric} employs symmetric transformations in conjunction with an Euler-Lagrangian formulation to approximate diffeomorphisms, while FireANTs~\cite{jena2024fireants} incorporates an adaptive Riemannian gradient descent algorithm to directly optimize transformations on manifolds.
Large Deformation Diffeomorphic Metric Mapping (LDDMM)~\cite{lddmm} leverages concepts from fluid mechanics and Lagrangian specifications to model substantial deformations. 
pTV~\cite{vishnevskiy2016isotropic}, based on B-splines, demonstrates the effectiveness of isotropic total variation alongside a linear grid interpolation strategy.
ConvexAdam~\cite{siebert2021fast} introduces an efficient approach that utilizes a feature extractor and correlation layer to initialize the deformation field, followed by Adam optimization for final DVF calculation.
Overall, while these classical methods achieve high registration accuracy through case-specific optimization, they tend to be computationally expensive when processing large images and require extensive hyperparameter tuning, which limits their scalability in clinical usage. In contrast, our work presents a case-specific optimization method that accelerates the registration process without compromising accuracy.

\subsection{Deep learning-based Registration} 

Inspired by the success of CNNs \jbhi{in the artificial intelligence community}, researchers have begun to explore their applications in DIR. VM~\cite{balakrishnan2019voxelmorph} utilizes a 3D U-Net architecture to directly predict DVFs, while probabilistic extensions~\cite{dalca2019unsupervised} incorporate stationary velocity fields (SVFs) and scaling-squaring techniques for diffeomorphic integration.
To address large deformations, LapIRN~\cite{mok2020large} employs a Laplacian pyramid network that integrates SVFs within the log-Euclidean framework, thereby improving diffeomorphism and minimizing the risk of local minima. 
SymNet~\cite{mok2020fast} ensures bidirectional consistency by introducing symmetric predictions of mid-time deformation fields. 
SynthMorph~\cite{hoffmann2021synthmorph} enhances generalization by generating diverse anatomical images for training.
To alleviate the challenges of hyperparameter tuning, HyperMorph~\cite{hoopes2022learning} adopts a meta-learning approach to dynamically adjust the hyperparameters of the registration network. 
ICON~\cite{greer2021icon} and its successors, GradICON~\cite{tian2023gradicon} and ConstrICON~\cite{greer2023inverse}, focus on inverse consistency for diffeomorphic registration, ensuring that transformations remain invertible. 
Emerging models such as DiffuseMorph~\cite{kim2022diffusemorph} and FSDiffReg~\cite{qin2023fsdiffreg} incorporate diffusion models to enhance deformation field prediction.
SDHNet~\cite{zhou2023self} utilizes a hierarchical network and introduces self-distillation to constrain intermediate DVFs. 
RClaNet~\cite{wu2023rclanet} employs a patch-based registration strategy to address the scarcity of training datasets and improve network performance in local regions. 
RegFSC-Net~\cite{liu2024regfsc} learns low-dimensional representations using an enhanced encoder and incorporates a parameter-free decoder that employs Fourier transformations to reduce computational costs.

Furthermore, Transformer architectures have demonstrated promising performance in various computer vision tasks by virtue of effectively modeling spatial long-range dependencies. 
Transmorph~\cite{chen2022transmorph} integrates Swin Transformer blocks within its encoder and presents both diffeomorphic and Bayesian variants. Later, Transmatch~\cite{chen2023transmatch} employs explicit feature matching between fixed and moving images by independently extracting features using self-attention and then fusing these features through cross-attention.
In summary, DL-based methods reduce processing time vastly, but they still struggle with generalization across diverse clinical scenarios~\cite{jena2024deep}.

\subsection{Implicit Neural Representation}
INR has recently demonstrated effectiveness in 3D scene reconstruction~\cite{mildenhall2021nerf}, with increasing applications in medical image registration. 
One pioneering INR-based method, IDIR~\cite{wolterink2022implicit}, employs a MLP to represent transformations between fixed and moving images. 
Unlike conventional CNN frameworks, IDIR is resolution-agnostic, allowing it to handle varying image resolutions without the need for resampling. 
Additionally, the tendency of traditional MLPs to learn low-frequency information enhances the smoothness of the deformation velocity field (DVF). 
To further improve diffeomorphism, ccIDIR~\cite{van2023robust} introduces dual cycle consistency loss, using each INR as a regularizer for the other, thereby stabilizing the optimization process. 
Despite INR’s advantages in flexibility and smoothness, the approach lacks interpretability due to its implicit nature, and it is still slower than required for real-time applications.

\subsection{3D Gaussian Splatting} 
The recent emergence of 3D GS\cite{kerbl20233d} has revolutionized 3D reconstruction in computer graphics, enabling real-time rendering of high-quality 3D scenes. 
Subsequent works have further enhanced rendering quality\cite{yu2024mip, fan2024trim, yan2024multi} and reduced computational burdens~\cite{lee2024compact}, while also generalizing to other fields such as occupancy prediction~\cite{huang2024gaussianformer}, dynamic scene reconstruction~\cite{huang2023sc, kratimenos2023dynmf}, and 2D image compression~\cite{zhang2024gaussianimage}. 
TrimGS~\cite{fan2024trim} focuses on improving adaptive density control, allowing for contribution-based trimming and scale-driven densification to eliminate redundant Gaussian primitives while maintaining relatively small Gaussian scales. 
MS3DGS~\cite{yan2024multi} incorporates multi-scale 3D Gaussian representations to address aliasing issues and further accelerate rendering. C3DGS~\cite{lee2024compact} employs vector quantization techniques to compress geometric attributes, including scaling and rotation.
GaussianFormer~\cite{huang2024gaussianformer} uses Gaussian primitives to represent the semantics of 3D space, deriving final occupancy predictions through the blending of truncated 3D primitives. 
DynMF~\cite{kratimenos2023dynmf} models the motion field as a linear combination of multiple neural trajectories predicted across all timestamps by a lightweight MLP.
Among these applications of GS, our proposed GaussianDIR is the first to lend the power of the Gaussian representation to the modeling of deformation field.
\section{Method}\label{sec:method}
The methodology is structured as follows: Sec.~\ref{sec:problem} presents the problem statement, followed by a detailed description of Gaussian primitives in Sec.~\ref{sec:GaussianPrimitive}. Next, Sec.~\ref{sec:blending} outlines the derivation of dense deformation fields through transformation blending. To enhance both efficiency and precision, adaptive and multi-scale Gaussian primitives are introduced in Sec.~\ref{sec:adaptive} and \ref{sec:hierarchy}, respectively. Finally, optimization and regularization strategies are discussed in Sec.~\ref{sec:optimization}. Fig.~\ref{fig:framework} provides a visual summary of the entire method, illustrating the sequential steps and interactions in our proposed approach.

\subsection{Problem Statement}\label{sec:problem}
The task addressed in this paper is pairwise image registration, focusing on finding the optimal spatial transformation between two images from the same domain. Specifically, we aim to calculate the displacement vector field (DVF), which represents the voxel-wise transformation required to align a moving image with a fixed image. Consider a pair of 3D images, $I_0$ and $I_1$, with voxels  $\boldsymbol{x} \in \mathbb{Z}^3$. Deformable image registration (DIR) estimates the DVF, denoted as $\boldsymbol{\phi}$, that maps each voxel of $I_0$ to its corresponding location in $I_1$. The optimization problem for DIR is formulated as follows:

\begin{equation} \label{eq:formulation}
\boldsymbol{\hat{\phi}} = \arg\min_{\boldsymbol{\phi}} \mathcal{L}(I_0 \circ \boldsymbol{\phi}, I_1) + \mathcal{R} (\boldsymbol{\phi}),
\end{equation}
where $I_1 \circ \boldsymbol{\phi}$ refers to the warped image $I_1$ after applying $\boldsymbol{\phi}$, and $\mathcal{L}$ measures the image similarity. The term $\mathcal{R}$ represents a regularization applied to the DVF $\boldsymbol{\phi}$ to constrain the solution space, avoiding singular solutions.

\subsection{Gaussian Primitives} \label{sec:GaussianPrimitive}
Our proposed approach models the deformation field using a set of sparsely distributed Gaussian primitives, offering an explicit solution in contrast to the implicit counterpart used in IDIR~\cite{wolterink2022implicit}. 
Each Gaussian primitive $\boldsymbol{G}_i$ is characterized by its center position $\boldsymbol{\mu}_i$ and covariance matrix $\boldsymbol{\Sigma}_i$, which is decomposed as $\boldsymbol{\Sigma}_i = \boldsymbol{Q}_i\boldsymbol{S}_i\boldsymbol{S}_i^T\boldsymbol{Q}_i^T$. Here, $\boldsymbol{S}_i$ is a scaling matrix parameterized by a diagonal vector $\boldsymbol{s}_i \in \mathbb{R}^3$, and $\boldsymbol{Q}_i \in \mathbf{SO}(3)$ is a rotation matrix represented by a quaternion $\boldsymbol{q}_i \in \mathbb{R}_i^4$. To describe the deformation, each Gaussian primitive is associated with a learnable local transformation $\boldsymbol{T}_i \in \mathbf{SE}(3)$, which is further decomposed into a rotation matrix $\boldsymbol{R}_i \in \mathbf{SO}(3)$, represented by a quaternion $\boldsymbol{r}_i \in \mathbb{R}^4$, and a translation vector $\boldsymbol{t}_i \in \mathbb{R}^3$. Consequently, the Gaussian primitives are parameterized as:


\begin{equation}
 \boldsymbol{G} = \{ \left(\boldsymbol{\mu}_i, \boldsymbol{s}_i, \boldsymbol{q}_i, \boldsymbol{r}_{i}, \boldsymbol{t}_i \right) \mid i \leq N\},
\end{equation}
where $N$ is the number of Gaussian primitives.

Overall, the sparse distribution of Gaussian primitives provides a compact basis for modeling the dynamic transformations occurring between images due to motion or morphological changes. This approach significantly reduces computational complexity while enhancing the framework's flexibility in capturing the essential dynamics of the transformation.

\subsection{$\alpha$-Transformation-Blending}\label{sec:blending}
The subsequent procedure is to calculate the dense deformation field for the volume. Inspiration by $\alpha$-blending~\cite{kerbl20233d}, we introduce the $\alpha$-transformation-blending, which determines the transformations of a voxel based on its surrounding Gaussian primitives. Specifically, we perform a K-nearest neighbor (KNN) search to identify $K$ closest Gaussian primitives of voxel $\boldsymbol{x}_j$. 
The blending weights $\boldsymbol{w}_{jk}$ for voxel $\boldsymbol{x}_j$ and Gaussian primitive $\boldsymbol{G}_k$ are calculated as such:
\begin{equation}\label{eq:weight}
    \boldsymbol{w}_{jk} = \frac{\boldsymbol{\hat{w}}_{jk}}{\sum\limits_{k\in\mathcal{N}_j}\boldsymbol{\hat{w}}_{jk}},
\end{equation}

\begin{equation}
    \boldsymbol{\hat{w}}_{jk} = \frac{1}{\sqrt{(2\pi)^3|\boldsymbol{\Sigma}_k|}}\exp(-\frac{1}{2}(\boldsymbol{x}_j - \boldsymbol{\mu}_k)^{T}\boldsymbol{\Sigma}_k^{-1}(\boldsymbol{x}_j - \boldsymbol{\mu}_k)),
\end{equation}
where $\mathcal{N}_j$ denotes the indices of the $K$ nearest Gaussian primitives of voxel $\boldsymbol{x}_j$.

Each Gaussian primitive impacts a local region in 3D volume, applying an aforementioned $\mathbf{SE}(3)$ transformation to the covered voxels. The transformed position component of voxel $\boldsymbol{x}_j$, contributed by $\boldsymbol{G}_k$, is denoted as $\boldsymbol{\hat{x}}_{jk}$, which is determined as follows:
\begin{equation} \label{eq:position}
    \boldsymbol{\hat{x}}_{jk} = \boldsymbol{R}_{k} (\boldsymbol{x}_{j} - \boldsymbol{\mu}_{k}) + \boldsymbol{\mu}_{k} + \boldsymbol{t}_{k}
\end{equation}

Given the normalized blending weights and transformed position components, we can calculate the displacement vector $\boldsymbol{\phi}_{j}$ as the weighted sum of the local rigid transformation $\boldsymbol{T}_{jk}$:

\begin{equation}\label{eq:linearcomb}
    \boldsymbol{\phi}_j = \sum\limits_{k\in\mathcal{N}_j}\boldsymbol{w}_{jk}\boldsymbol{T}_{jk},\quad
    \boldsymbol{T}_{jk} = \boldsymbol{\hat{x}}_{jk} - \boldsymbol{x}_{j}
\end{equation}

This blending method enables smooth transitions between Gaussian primitives, effectively capturing the continuous nature of biological deformations. Additionally, by adopting this strategy, similar to the truncated Gaussian approach in 3D GS, our method significantly reduces computational complexity,
thus allows for a more significant number of Gaussian primitives, facilitating the modeling of more complex deformations.

\begin{figure}[htbp]
    \centering
    \includegraphics[width=0.98\linewidth]{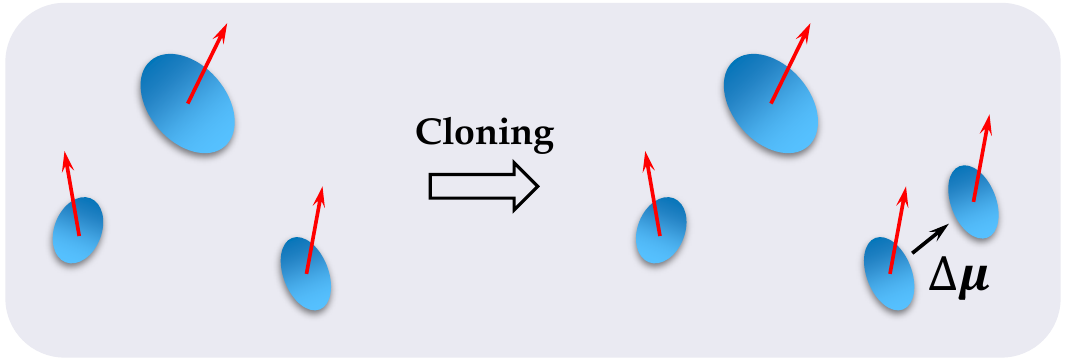}
    \caption{Gaussian primitives are densified by cloning those whose position gradients exceed the threshold $\boldsymbol{\tau}_{\text{max}}$, with a random displacement $\boldsymbol{\Delta\mu}$ applied to the clones.}
    \label{fig:dp_clone}
\end{figure}

\subsection{Adaptive Density} \label{sec:adaptive}
The density of Gaussian primitives is a critical hyper-parameter, directly influencing the complexity of the DVF modeling. Striking a delicate balance is critical: excessively dense Gaussian primitives can lead to over-fitting to noise, thereby compromising the smoothness of the resultant deformation field. Conversely, too sparse Gaussian primitives may result in under-fitting, particularly in regions with complex deformations. Manually adjusting this hyper-parameter for each pair of images is both inefficient and impractical.

To address this challenge, we employ an adaptive density control scheme inspired by previous works~\cite{kerbl20233d, huang2023sc}, where the density of Gaussian primitives is dynamically adjusted. 
Specifically, we determine whether a Gaussian primitive should be cloned or pruned based on the norm of its position gradient norm. Gaussian primitives with gradient norm exceeding a predefined threshold $\boldsymbol{\tau}_{\text{max}}$ are cloned, while those with norm below $\boldsymbol{\tau}_{\text{min}}$ are pruned.
The cloning process, illustrated in Fig.~\ref{fig:dp_clone}, involves duplicating a Gaussian primitive $\boldsymbol{G}_i$ by sampling a random displacement $\Delta\boldsymbol{\mu}_i$ from a 3D normal distribution $\boldsymbol{N}(\boldsymbol{0}, \boldsymbol{s}_i)$, where the $\boldsymbol{s}_i$ denotes the scaling vector of $\boldsymbol{G}_i$.
The coordinate $\boldsymbol{\hat{\mu}}_i$ of the new Gaussian is determined by adding this random displacement to the original position as $\boldsymbol{\hat{\mu}}_i = \boldsymbol{\mu}_i + \Delta\boldsymbol{\mu}_i$, while the other parameters are retained. 
The gradient of new Gaussian primitives are initially set as zeros and will be calculated though back-propagation in the next iteration.
Pruning is achieved by straightforwardly removing the corresponding Gaussian primitives.

\subsection{Multi-scale Gaussian Primitives} \label{sec:hierarchy}
The multi-scale scheme is commonly employed to mitigate the impact of local minima, either through image pyramids~\cite{bajcsy1989multiresolution, vishnevskiy2016isotropic} or pyramid networks~\cite{zhou2023self, wang2024recursive}. 
In our framework, we progressively refine the deformation field across multiple-scale Gaussian primitives without downsampling the original images. 
This is achieved by initializing Gaussian primitives at multiple levels of granularity, with varying maximum numbers of primitives across different scales.
The number of Gaussian primitives directly influences the complexity of the deformation field, allowing for a coarse-to-fine refinement of the DVF.

As illustrated in Fig.~\ref{fig:framework}(c), we begin by estimating a coarse deformation field $\boldsymbol{\phi}^c$ using larger, sparser Gaussian primitives. This is followed by a finer deformation field $\boldsymbol{\phi}^f$ derived from smaller, denser Gaussian primitives. The final deformation field $\boldsymbol{\phi}$ is obtained by combining these fields: $\boldsymbol{\phi} = \boldsymbol{\phi}^c + \boldsymbol{\phi}^f$. This multi-scale facilitates reducing computational complexity while maintaining precision.

\subsection{Optimization}\label{sec:optimization}
We employ the mini-batch gradient descent technique to optimize the parameters of the Gaussian primitives, similar to the strategy used in IDIR~\cite{wolterink2022implicit}. The normalized cross-correlation (NCC)\cite{rao2014application}, defined in Eq.\ref{eq:NCC}, serves as the similarity measure to guide the optimization process.

\begin{equation}\label{eq:NCC}
    \scalebox{1.1}{$
    \mathcal{L}_s(I_f, I_w) = -\frac{\sum_{x_i} (I_f(x_i) - \bar{I}_f)(I_w(x_i) - \bar{I}_w)}{\sqrt{\sum_{x_i} (I_f(x_i) - \bar{I}_f)^2 \sum_{x_i} (I_w(x_i) - \bar{I}_w)^2}},
    $}
\end{equation}
where $\bar{I}_f$ and $\bar{I}_w$ represent the mean intensity of the fixed image $I_f$ and the warped image $I_w$, respectively, and $x_i$ denotes the voxel location in the image. The NCC loss is chosen for its robustness to intensity variations and its ability to maintain structural integrity during registration.


In addition to the similarity measure, we incorporate the total variation (TV) regularization~\cite{vishnevskiy2016isotropic}, denoted as $\mathcal{L}_r$, to promote the smoothness of the DVF. The TV regularization penalizes the spatial gradients of the DVF, thereby ensuring smooth transitions in the deformation.

However, traditional TV regularization is designed for whole images and is incompatible with our mini-batch-based optimization strategy. To address this, we develop an efficient mini-batch-based TV regularization method. Specifically, in each iteration, we randomly sample $\frac{1}{D+1}B$ voxels, where $B$ is the batch size and $D$ is the image dimension. For each sampled voxel, we also include its neighboring voxels in three orthogonal directions to compute the TV loss. This sampling strategy reduces computational overhead while preserving the regularization effect. A 2D illustration of the TV regularization method is shown in Fig.~\ref{fig:framework}(b).

The final loss function is thus expressed as: \begin{equation}
    \mathcal{L} = \mathcal{L}_s + \lambda\mathcal{L}_r, 
\end{equation}
where $\lambda$ is the weight balancing the regularization term.

\section{Experiments}\label{sec:experiments}
\subsection{Datasets}
We conduct experiments on diverse datasets, including brain MRI, lung CT, and cardiac MRI images, which are significant for advancing medical DIR. 
These datasets cover multiple clinical scenarios, enhancing the applicability of our methods in real-world settings like neurodegenerative disease monitoring, cardiac motion analysis, and respiratory analysis. The specific datasets are as follows:
\paragraph{OASIS}

We utilize the Neurite version~\cite{hoopes2021hypermorph} of the OASIS dataset~\cite{marcus2007open} for brain MRI evaluation, which includes $414$ subjects with $35$ anatomical labels. To ensure comprehensive comparisons with benchmark methods, we use two subsets: (\romannumeral1) Following \cite{jena2024deep}, we select the last $50$ subjects to form $49$ test pairs, adhering to the pairing rules from the Learn2Reg challenge. (\romannumeral2) In line with SDHNet~\cite{zhou2023self}, we choose the last $20$ subjects to generate $20 \times 19$ test pairs for pairwise registration.

\paragraph{DIR-Lab} 
For lung CT evaluation, we adopt the DIR-Lab dataset~\cite{castillo2009framework}, which consists of $10$ lung 4D CT images, each annotated with $300$ anatomical landmarks. 
We extract the extreme inhalation and exhalation images as fixed and moving images, respectively, preserving the original resolution and unnormalized intensity values.

\paragraph{ACDC} 

We use the ACDC dataset~\cite{bernard2018deep} for cardiac MRI registration, which contains $150$ image pairs captured between the end-diastolic (ED) and end-systolic (ES) phases. Our test set and preprocessing procedures align with those used in CorrMLP~\cite{meng2024correlation}.

\paragraph{IXI}
The IXI dataset~\cite{ixidataset}, preprocessed by Transmorph~\cite{chen2022transmorph}, is employed to compare the generalization of our method with DL-based approaches. We follow the Transmorph protocol, using the provided test set and atlas-based registration to assess registration accuracy.

\subsection{Metrics}\label{metrics}
We adopt standard evaluation metrics from prior works~\cite{vishnevskiy2016isotropic, wolterink2022implicit, balakrishnan2019voxelmorph, zhou2023self, meng2024correlation}: target registration error (TRE), dice similarity coefficient (DSC), HD95, and negative Jacobian determinant (NJD). TRE measures the mean Euclidean distance between anatomical landmarks in the fixed and warped moving images, making it suitable for DIRLab. DSC evaluates the overlap between segmented regions. HD95 calculates the $95\%$ maximum Hausdorff distance between corresponding point sets, while NJD assesses the smoothness and physical plausibility of the deformation field.

\subsection{Implementation Details}
We initialize the positions of the Gaussian primitives at nodes of a 3D grid within Pytorch's canonical space $[-1, 1]^3$, while the local rigid transformations are set to the identity transformation. To expedite optimization, we employ a mini-batch optimization strategy with a batch size of $M = 20,000$ voxels, sampled from the 3D volume. The Adam optimizer is used for $2000$ iterations, with an initial warm-up phase for the learning rate, followed by a decay using a cosine annealing schedule.
To fully leverage the hierarchical scheme, we divide the optimization into two stages. In the first stage, we set the maximum number of Gaussian primitives to $\frac{1}{8} \times |\Omega|$, while in the second stage, the number is increased to $\frac{1}{4} \times |\Omega|$, starting at the $0.5 \times M$ iteration. The weight $\lambda$ for the regularization term is empirically set to $15$.
Additionally, we perform cloning and pruning of Gaussian primitives based on gradient norms. The cloning threshold $\boldsymbol{\tau}_{\text{max}}$ is set to $0.002$, while the pruning threshold $\boldsymbol{\tau}_{\text{min}}$ is set to $1 \times 10^{-7}$. These operations are conducted every $0.05 \times M$ iterations. All experiments are run on an Nvidia RTX 4090 GPU.

\begin{figure}
    \centering
    \includegraphics[width=\linewidth]{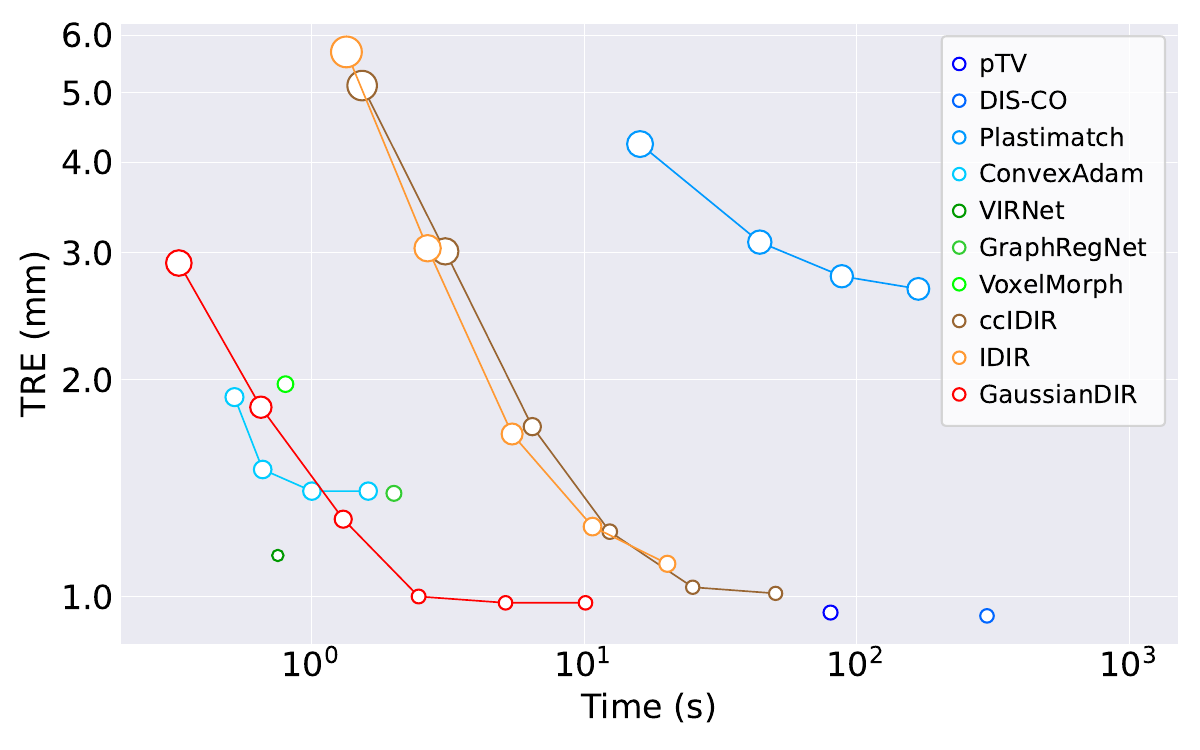}
    \caption{This line plot illustrates landmark error on the DIRLab dataset averaged across 10 cases with standard deviation represented by the size of the circles. Larger circles indicate greater standard deviations, reflecting higher variability in the data. Notably, GaussianDIR achieves the best trade-off between speed and accuracy.}
    \label{fig:DIRLab-time-tre}
\end{figure}

\begin{table}[htbp]
    \caption{Mean (standard deviation) DSC, HD95 and NJD on OASIS dataset with 49 pairs. The best results are bold.}
    \label{tab:OASIS-metric-49}
    \centering
    \resizebox{1.0\linewidth}{!}{
    \begin{tabular}{l|l|ccc}
    \toprule
    Methods   
    & Type     & DSC (\%) $\uparrow$  & HD95 $\downarrow$ & NJD (\%) $\downarrow$ \\
    \midrule
    NiftyReg~\cite{modat2010fast}
    & Classical & 77.6 (3.1) & 2.33 (0.76)  & 7.596 (3.14)\\
    SyN~\cite{avants2008symmetric}
    & Classical & 78.7 (3.1) & 2.16 (0.55)  & 0.008 (0.01)\\
    FireANTs~\cite{jena2024fireants}
    & Classical & 80.5 (2.9) & 1.96 (0.49)  & \textbf{0} \\
    Greedy~\cite{yushkevich2016ic}
    & Classical & 80.6 (2.9) & 1.94 (0.51)  & \textbf{0} \\
    Demons~\cite{vercauteren2009diffeomorphic}
    & Classical & 80.2 (2.4) & 2.02 (0.46)  & 0.094 (0.03) \\ 
    SymNet~\cite{mok2020fast}
    & DL      & 74.8 (-)   & - & -\\
    LapIRN~\cite{mok2020large}
    & DL      & 78.8 (-)   & - & -\\
    NODEO~\cite{wu2022nodeo}
    & INR      & 79.0 (2.7) & 2.05 (0.48) & 0.013 (0.01)\\
    IDIR~\cite{wolterink2022implicit}
    & INR      & 78.1 (2.6) & 2.25 (0.62) & 1.191 (0.21)\\
    ccIDIR~\cite{van2023robust}  
    & INR      & 78.4 (2.6) & 2.16 (0.58) & 0.763 (0.13)\\
    \midrule
    GaussianDIR
    & Gaussian & \textbf{81.3 (2.3)} & \textbf{1.89 (0.50)} & 1.091 (0.24) \\
    \bottomrule
    \end{tabular}}
\end{table}

\begin{table}[htbp]
    \caption{Mean (standard deviation) DSC, HD95 and NJD on OASIS dataset with 380 pairs. The best results are highlighted as bold. *: $P<0.05$, in comparison to GaussianDIR.}
    \label{tab:OASIS-metric}
    \centering
    \resizebox{1.0\linewidth}{!}{
    \begin{tabular}{l|l|ccc}
    \toprule
    Methods   & Type    
    & DSC (\%) $\uparrow$ & HD95 $\downarrow$ & NJD (\%) $\downarrow$ \\
    \midrule
    NiftyReg~\cite{modat2010fast}
    & Classical & 78.0 (2.8)* & 2.24 (0.54) & 8.512 (3.646)\\
    FireANTs~\cite{jena2024fireants}
    & Classical & 80.8 (2.8)* & 1.96 (0.44) & \textbf{0} \\
    Greedy~\cite{yushkevich2016ic}
    & Classical & 80.9 (2.8)* & 1.93 (0.45) & \textbf{0} \\
    Demons~\cite{vercauteren2009diffeomorphic}
    & Classical & 80.5 (2.2)* & 2.00 (0.37) & 0.095 (0.03) \\ 
    NODEO~\cite{wu2022nodeo}
    & INR       & 79.3 (2.9)* & 2.07 (0.47) & 0.014 (0.01)\\
    IDIR~\cite{wolterink2022implicit}
    & INR       & 78.9 (2.5)* & 2.21 (0.48) & 1.184 (0.23) \\
    ccIDIR~\cite{van2023robust}  
    & INR       & 79.0 (2.4)* & 2.13 (0.46) & 0.753 (0.14) \\
    \midrule
    GaussianDIR
    & Gaussian & \textbf{81.6 (2.1)} & \textbf{1.86 (0.39)} & 1.106 (0.22) \\
    \bottomrule
    \end{tabular}}
\end{table}

\begin{table*}[htbp]
    \caption{Evaluation of TRE ($\downarrow$) in \textit{mm} on the DIR-Lab dataset. Results for GaussianDIR are averaged over $5$ random seeds. The best results are highlighted as \colorbox{red!30}{first}, \colorbox{orange!30}{second}. *: $P<0.05$, in comparison to GaussianDIR.}
    \label{tab:DIRLab-metric}
    \centering
    \resizebox{1.0\textwidth}{!}{
    \begin{tabular}{c|ccc|ccc|ccccc}
    \toprule
    Methods   & GaussianDIR      
              & ccIDIR*~\cite{van2023robust} 
              & IDIR*~\cite{wolterink2022implicit} 
              & VIRNet*~\cite{hering2021cnn}
              & GraphNet*~\cite{ashburner2011diffeomorphic}
              & VM*~\cite{balakrishnan2019voxelmorph}
              & pTV~\cite{vishnevskiy2016isotropic}
              & ConvexAdam*~\cite{siebert2021fast}
              & Demons*~\cite{vercauteren2009diffeomorphic}
              & FireANTs*~\cite{jena2024fireants}
              & Greedy*~\cite{yushkevich2016ic}\\
    \midrule
    4DCT 01   & 0.78 (0.92) & 0.83 (0.94) & 0.76 (0.94) & 0.99 (0.47) & 0.86 (0.91)
              & 1.03 (1.01) & 0.76 (0.90) & 1.00 (1.03)
              & 1.07 (0.50) & 0.92 (0.96) & 0.85 (0.92)\\
    4DCT 02   & 0.75 (0.90) & 0.78 (0.93) & 0.76 (0.94) & 0.98 (0.46) & 0.90 (0.95)
              & 1.09 (1.87) & 0.77 (0.89) & 0.74 (0.96)
              & 1.14 (0.97) & 0.89 (0.96) & 0.94 (0.95)\\
    4DCT 03   & 0.94 (1.06) & 1.02 (1.10) & 0.94 (1.02) & 1.11 (0.61) & 1.06 (1.10)
              & 1.40 (2.04) & 0.90 (1.05) & 1.02 (1.11)
              & 1.32 (0.80) & 1.09 (1.08) & 1.05 (1.09)\\
    4DCT 04   & 1.25 (1.24) & 1.37 (1.36) & 1.32 (1.27) & 1.37 (1.03) & 1.45 (1.24)
              & 1.69 (2.60) & 1.24 (1.29) & 1.39 (1.35)
              & 1.86 (1.67) & 1.54 (1.47) & 1.35 (1.25)\\
    4DCT 05   & 1.11 (1.46) & 1.25 (1.51) & 1.23 (1.47) & 1.32 (1.36) & 1.60 (1.50)
              & 1.63 (2.44) & 1.12 (1.44) & 1.42 (1.63)
              & 1.75 (1.65) & 1.42 (1.57) & 1.39 (1.60)\\
    4DCT 06   & 0.96 (1.00) & 1.06 (1.09) & 1.09 (1.03) & 1.15 (1.12) & 1.59 (1.06)
              & 1.60 (2.58) & 0.85 (0.89) & 1.31 (1.25)
              & 2.33 (2.82) & 2.24 (3.14) & 3.12 (4.66)\\
    4DCT 07   & 0.93 (0.96) & 0.97 (0.98) & 1.12 (1.00) & 1.05 (0.81) & 1.74 (1.10)
              & 1.93 (2.80) & 0.80 (1.28) & 1.49 (1.73)
              & 4.18 (4.84) & 3.68 (5.57) & 4.30 (6.24)\\
    4DCT 08   & 1.10 (1.26) & 1.13 (1.40) & 1.21 (1.29) & 1.22 (1.44) & 1.46 (1.27)
              & 3.16 (4.69) & 1.34 (1.93) & 2.91 (5.13)
              & 6.92 (8.77) & 7.17 (9.50) & 8.77 (10.52)\\
    4DCT 09   & 1.00 (0.95) & 1.02 (0.93) & 1.22 (0.95) & 1.11 (0.66) & 1.58 (1.07)
              & 1.95 (2.37) & 0.92 (0.94) & 1.33 (1.32)
              & 1.56 (1.18) & 1.57 (1.92) & 1.81 (2.26)\\
    4DCT 10   & 0.92 (0.90) & 0.96 (0.98) & 1.01 (1.05) & 1.05 (0.72) & 1.71 (2.03)
              & 1.66 (2.87) & 0.82 (0.89) & 1.41 (2.11)
              & 2.43 (3.24) & 1.94 (3.40) & 2.43 (4.07)\\
    \midrule
    Average   & \cellcolor{orange!30}{0.97 (1.07)} & 1.04 (1.12) & 1.07 (1.10) & 1.14 (0.76) & 1.39 (1.29)
              & 1.71 (2.86) & \cellcolor{red!30}{0.95 (1.15)} & 1.40 (1.76)
              & 2.45 (1.72) & 2.24 (1.82) & 2.60 (2.30) \\
    \bottomrule
    \end{tabular}}
\end{table*}

\begin{table}[htbp]
  \caption{Mean DSC and NJD on the ACDC dataset. The best results are highlighted as bold.}
  \label{tab:ACDC-metric}
  \centering
  \begin{tabular}{l|c|cc}
    \toprule
    Methods 
    & Type & DSC (\%) $\uparrow$ & NJD (\%) $\downarrow$ \\
    \midrule
    Demons~\cite{vercauteren2009diffeomorphic}
    & Classical & 79.4 & 3.400 \\
    FireANTs~\cite{jena2024fireants}
    & Classical & 79.5 & 0.003 \\
    Greedy~\cite{yushkevich2016ic}
    & Classical & 80.0 & \textbf{0.002} \\
    VM~\cite{balakrishnan2019voxelmorph}
    & DL & 75.4 & 0.440 \\
    TransMorph~\cite{chen2022transmorph}
    & DL & 76.8 & 0.492 \\
    TransMatch~\cite{chen2023transmatch}
    & DL & 77.0 & 0.425 \\
    LapIRN~\cite{mok2020large}
    & DL & 79.0 & 0.454 \\
    Dual-PRNet++~\cite{kang2022dual}
    & DL & 77.7 & 0.479 \\
    SDHNet~\cite{zhou2023self}
    & DL & 78.9 & 0.395 \\
    CorrMLP~\cite{meng2024correlation}
    & DL & 81.0 & 0.389 \\
    NODEO~\cite{wu2022nodeo}
    & INR & 80.3 & 0.008 \\
    IDIR~\cite{wolterink2022implicit}
    & INR & 80.6 & 0.496 \\
    ccIDIR~\cite{van2023robust}
    & INR & 80.0 & 0.113 \\
    \midrule
    GaussianDIR
    & Gaussian & \textbf{81.2} & 0.347 \\
    \bottomrule
  \end{tabular}
\end{table}

\subsection{Results}
We conduct a comprehensive comparative analysis against a broad range of baseline methods, including classical registration methods such as Demons\cite{vercauteren2009diffeomorphic}, NiftyReg\cite{modat2010fast}, pTV\cite{vishnevskiy2016isotropic}, Greedy\cite{yushkevich2016ic} and ConvexAdam\cite{siebert2021fast}, FireANTs\cite{jena2024fireants}, learning-based methods including VM\cite{balakrishnan2019voxelmorph}, SymNet\cite{mok2020fast}, and LapIRN\cite{mok2020large}, GraphNet\cite{hansen2021graphregnet}, VIRNet\cite{hering2021cnn}, Dual-PRNet++\cite{kang2022dual}, TransMorph\cite{chen2022transmorph}, SDHNet\cite{zhou2023self}, TransMatch\cite{chen2023transmatch}, CorrMLP\cite{meng2024correlation}, and INR-based methods like NODEO\cite{wu2022nodeo}, IDIR\cite{wolterink2022implicit}, ccIDIR\cite{van2023robust}. These methods are all highly cited and influential in the field.

\begin{figure*}
    \centering
    \includegraphics[width=\linewidth]{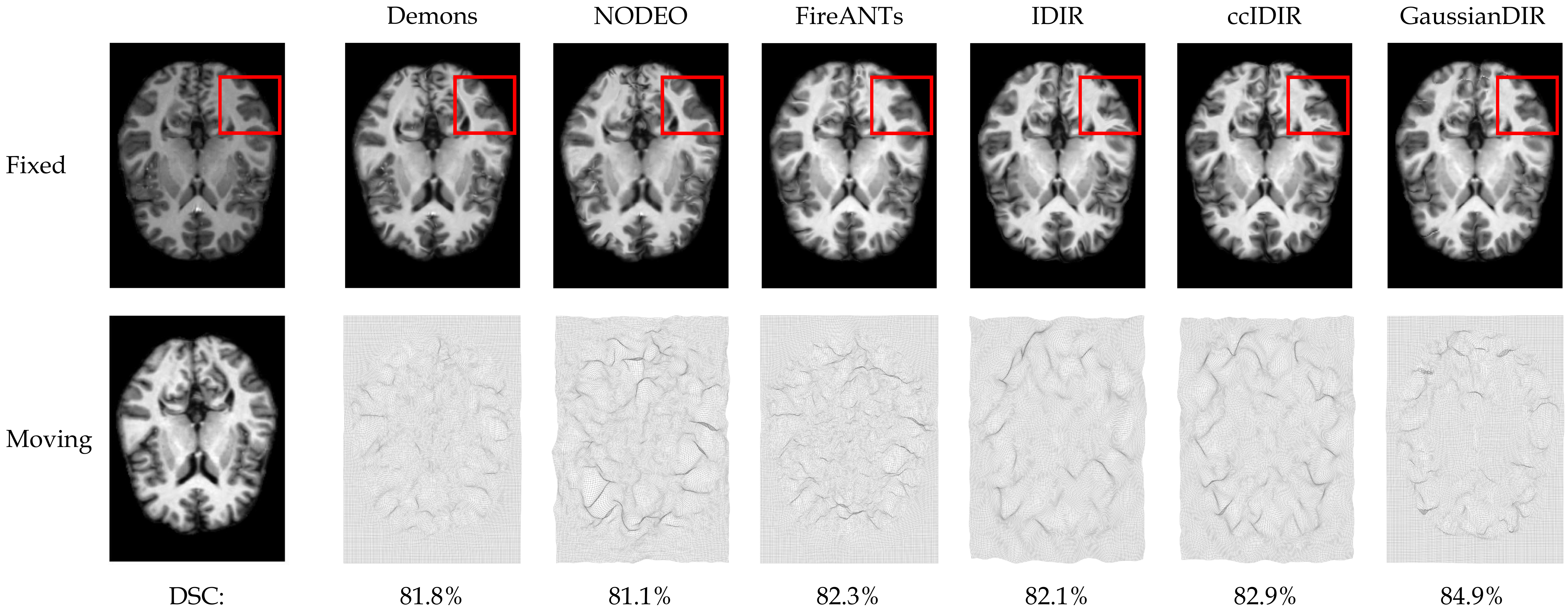}
    \caption{Visualization of warped moving images and deformation field on the OASIS dataset. Their corresponding DSC values are presented in the bottom row.}
    \label{fig:vis_OASIS}
\end{figure*}

\paragraph{Brain MRI}
Tab.~\ref{tab:OASIS-metric-49} demonstrate that GaussianDIR outperforms classical, DL-based, INR-based methods on the OASIS dataset with $49$ pairs. It achieves a higher DSC of $81.3\pm2.3\%$ and a lower HD95 of $1.89\pm0.50$. 
As shown in Tab.~\ref{tab:OASIS-metric}, when applied to the OASIS dataset with $380$ pairs, GaussianDIR maintains robust performance, achieving a DSC of $81.6\pm2.1\%$ and a HD95 of $1.86\pm0.39\%$. 
Fig.~\ref{fig:vis_OASIS} illustrates comparisons of warped moving images and the corresponding DVFs. The highlighted regions within the red boxes indicate that the warped moving images generated by GaussianDIR align more closely with the fixed images. Furthermore, the DVFs highlight GaussianDIR's strong edge-preserving capabilities, which are crucial for maintaining anatomical consistency.

\paragraph{Lung CT}
Detailed case-wise TRE comparisons on the DIR-Lab dataset are presented in Table~\ref{tab:DIRLab-metric}. Notably, GaussianDIR surpasses most methods in TRE, with the exception of pTV~\cite{vishnevskiy2016isotropic}. However, our approach offers significantly faster computation times compared to pTV, as evidenced in Fig.~\ref{fig:DIRLab-time-tre}, which shows that GaussianDIR achieves an optimal balance between accuracy and processing time.

\paragraph{Cardiac MRI}
Tab.~\ref{tab:ACDC-metric} summarizes the results on the ACDC cardiac dataset. GaussianDIR achieves the highest DSC of $81.2\%$, demonstrating the method's capability to generalize effectively across different types of DIR tasks.

\paragraph{Statistical Significance}
We perform statistical significance tests using the DSC results from the OASIS dataset and the TRE results from the DIRLab dataset. A p-value less than 0.05 indicates a statistically significant difference between methods. In Tab.~\ref{tab:OASIS-metric} and \ref{tab:DIRLab-metric}, we highlight methods that exhibit significant differences compared to GaussianDIR. As shown, GaussianDIR differs significantly from all other methods except for pTV~\cite{vishnevskiy2016isotropic} (p = 0.57), demonstrating that GaussianDIR significantly outperforms most methods while showing no substantial registration degradation compared to pTV.

\begin{figure*}[htbp]    
    \centering
    \begin{subfigure}[b]{0.24\textwidth}
        \centering
        \includegraphics[width=\textwidth]{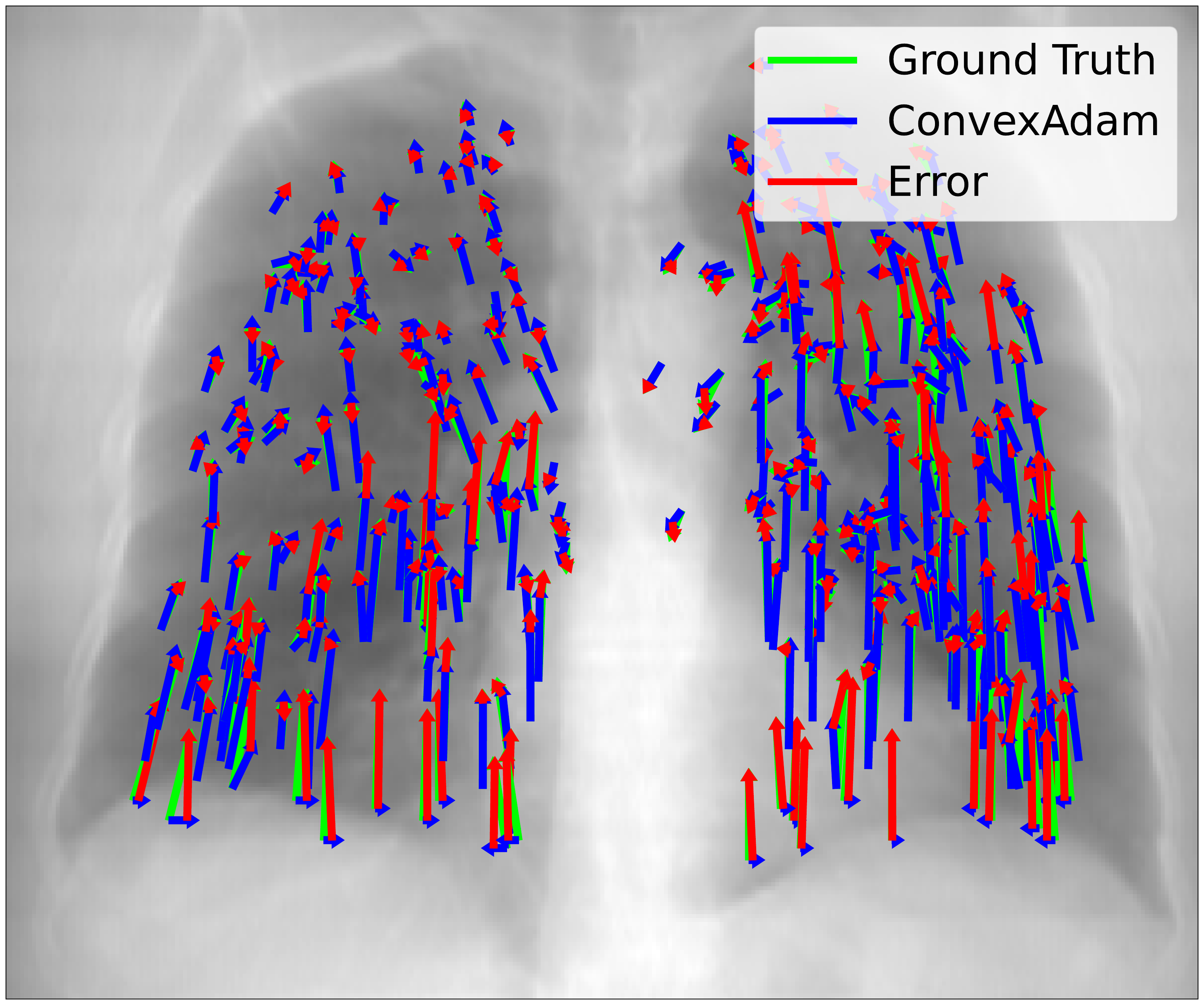}
        \caption{ConvexAdam~\cite{siebert2021fast}}
    \end{subfigure}
    \hfill
    \begin{subfigure}[b]{0.24\textwidth}
        \centering
        \includegraphics[width=\textwidth]{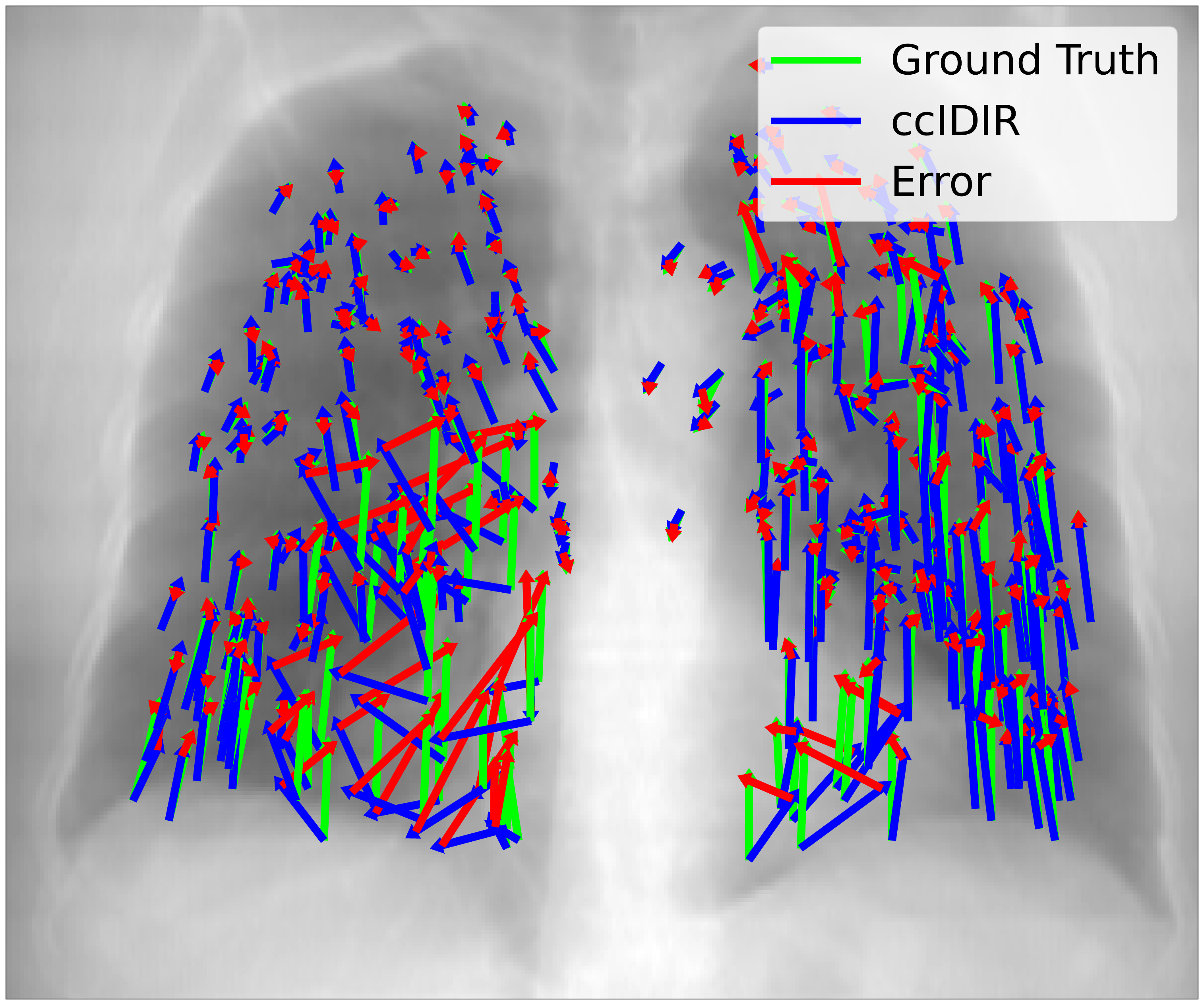}
        \caption{ccIDIR~\cite{van2023robust}}
    \end{subfigure}
    \hfill
    \begin{subfigure}[b]{0.24\textwidth}
        \centering
        \includegraphics[width=\textwidth]{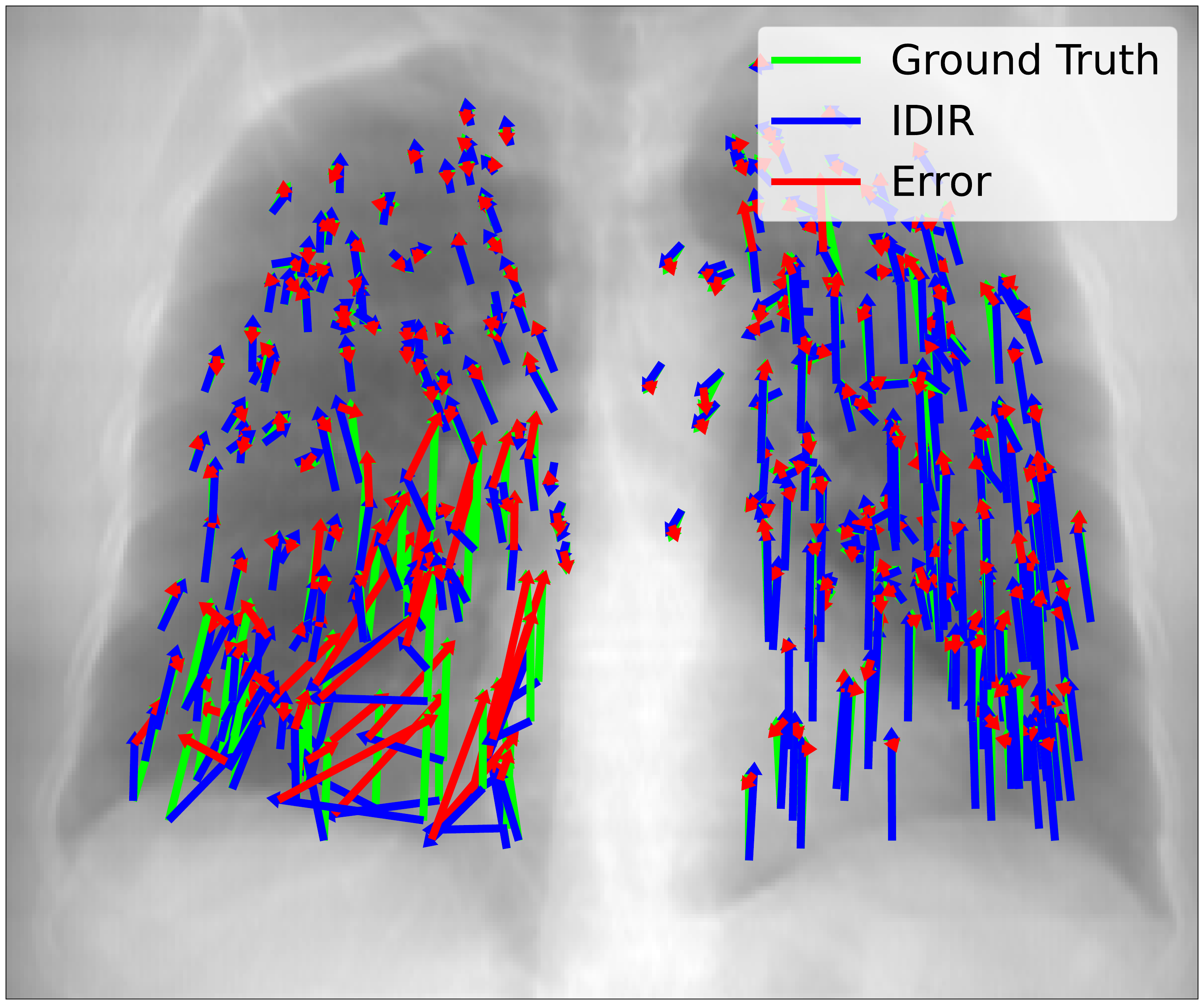}
        \caption{IDIR~\cite{wolterink2022implicit}}
    \end{subfigure}
    \hfill
    \begin{subfigure}[b]{0.24\linewidth}
        \centering
        \includegraphics[width=\linewidth]{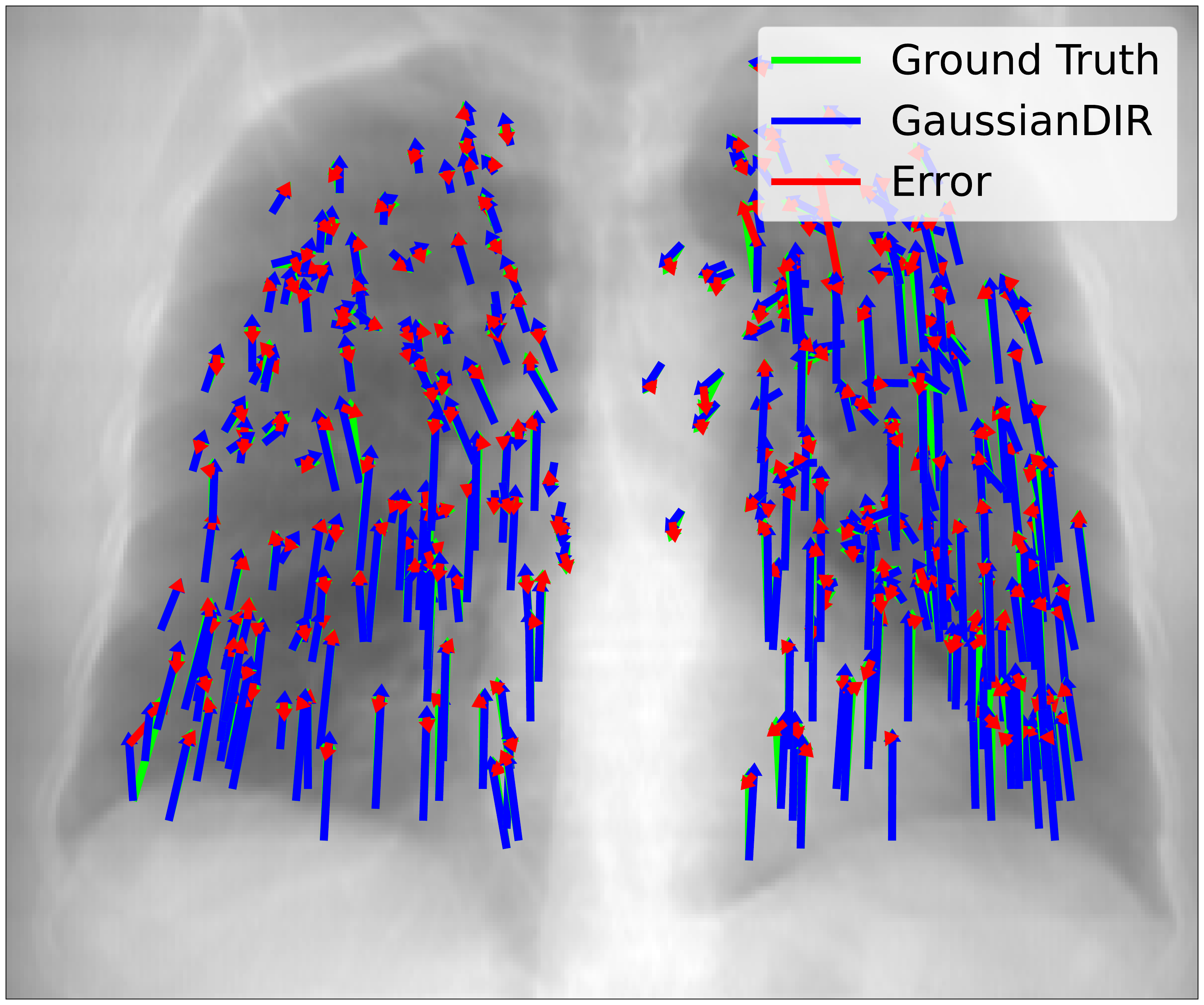}
        \caption{GaussianDIR.}
    \end{subfigure}
    \caption{Landmark motion visualization for Case 8 of DIRLab dataset. The visualization compares four methods each trained for approximately 2.5 seconds. Blue arrows represent predicted motion, green arrows denote ground-truth motion, and red arrows indicate error. Shorter red arrows correspond to smaller errors, demonstrating more accurate predictions.}
    \label{fig:DIRLabtrack}
\end{figure*}

\begin{figure*}[htbp]
    \centering
    \begin{subfigure}[b]{0.19\textwidth}
        \centering
        \includegraphics[width=\textwidth]{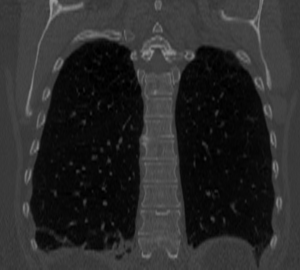}
        \caption{Fixed Image}
    \end{subfigure}
    \hfill
    \begin{subfigure}[b]{0.19\textwidth}
        \centering
        \includegraphics[width=\textwidth]{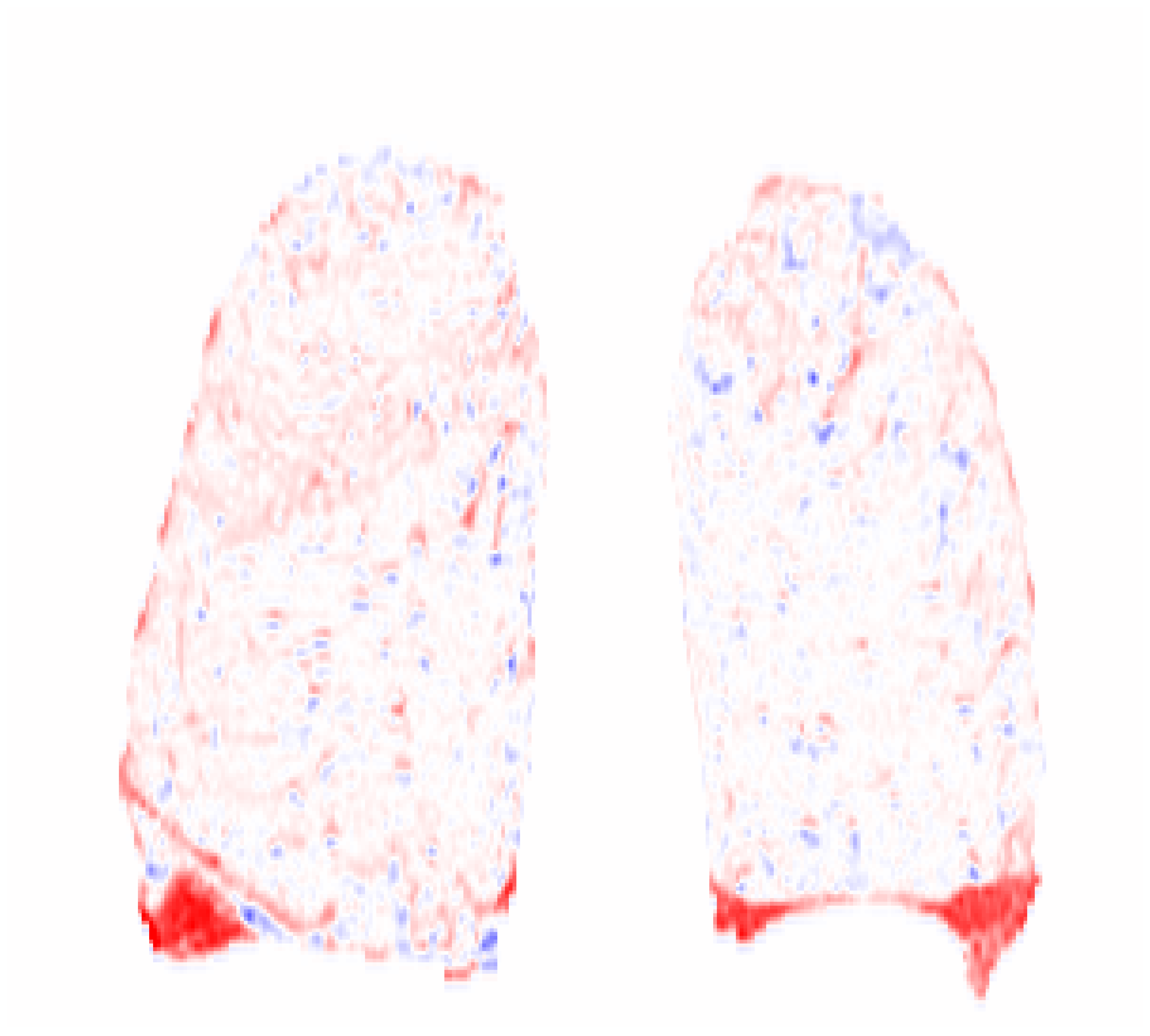}
        \caption{ConvexAdam~\cite{siebert2021fast}}
    \end{subfigure}
    \hfill
    \begin{subfigure}[b]{0.19\textwidth}
        \centering
        \includegraphics[width=\textwidth]{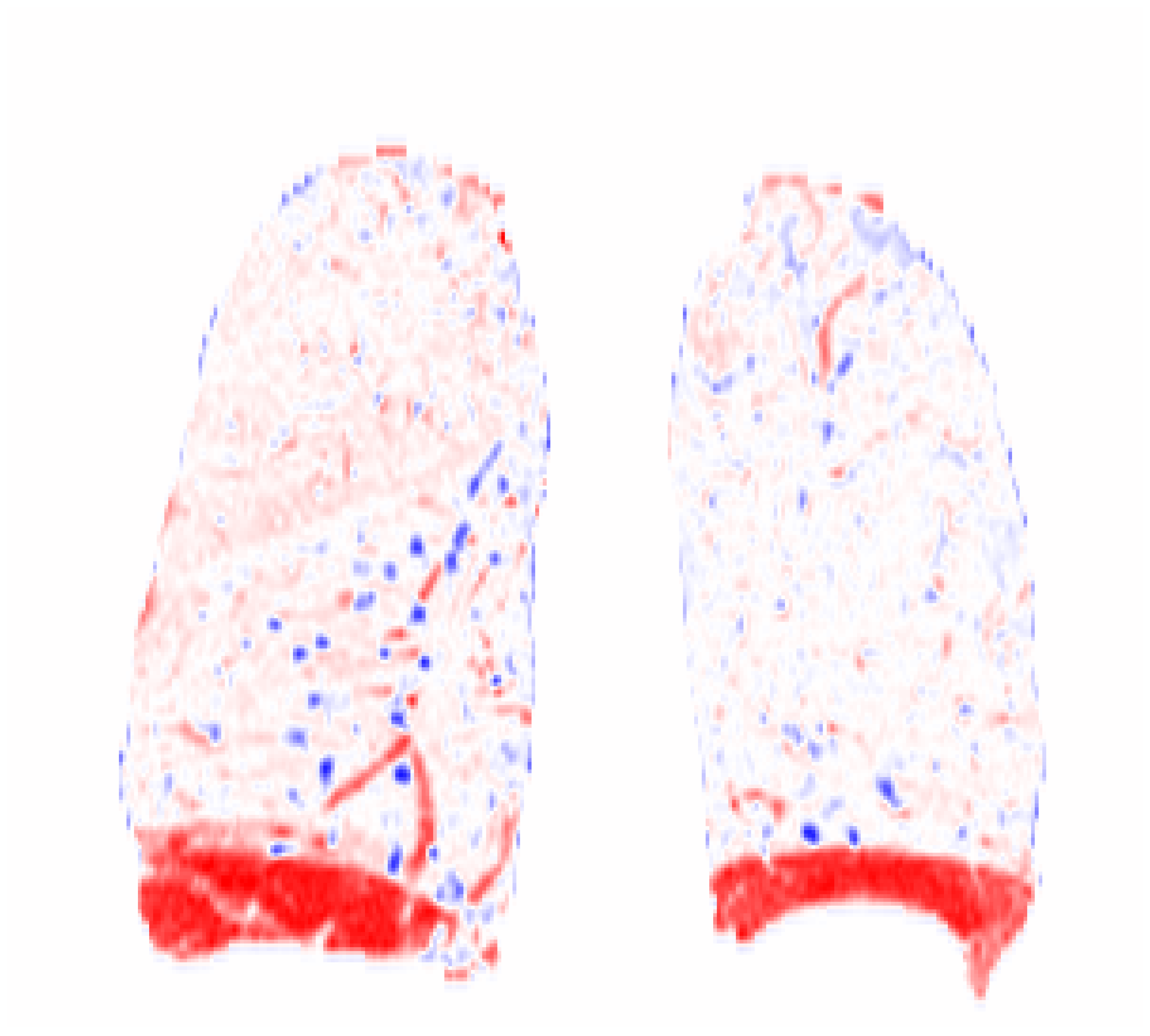}
        \caption{ccIDIR~\cite{van2023robust}}
    \end{subfigure}
    \hfill
    \begin{subfigure}[b]{0.19\textwidth}
        \centering
        \includegraphics[width=\textwidth]{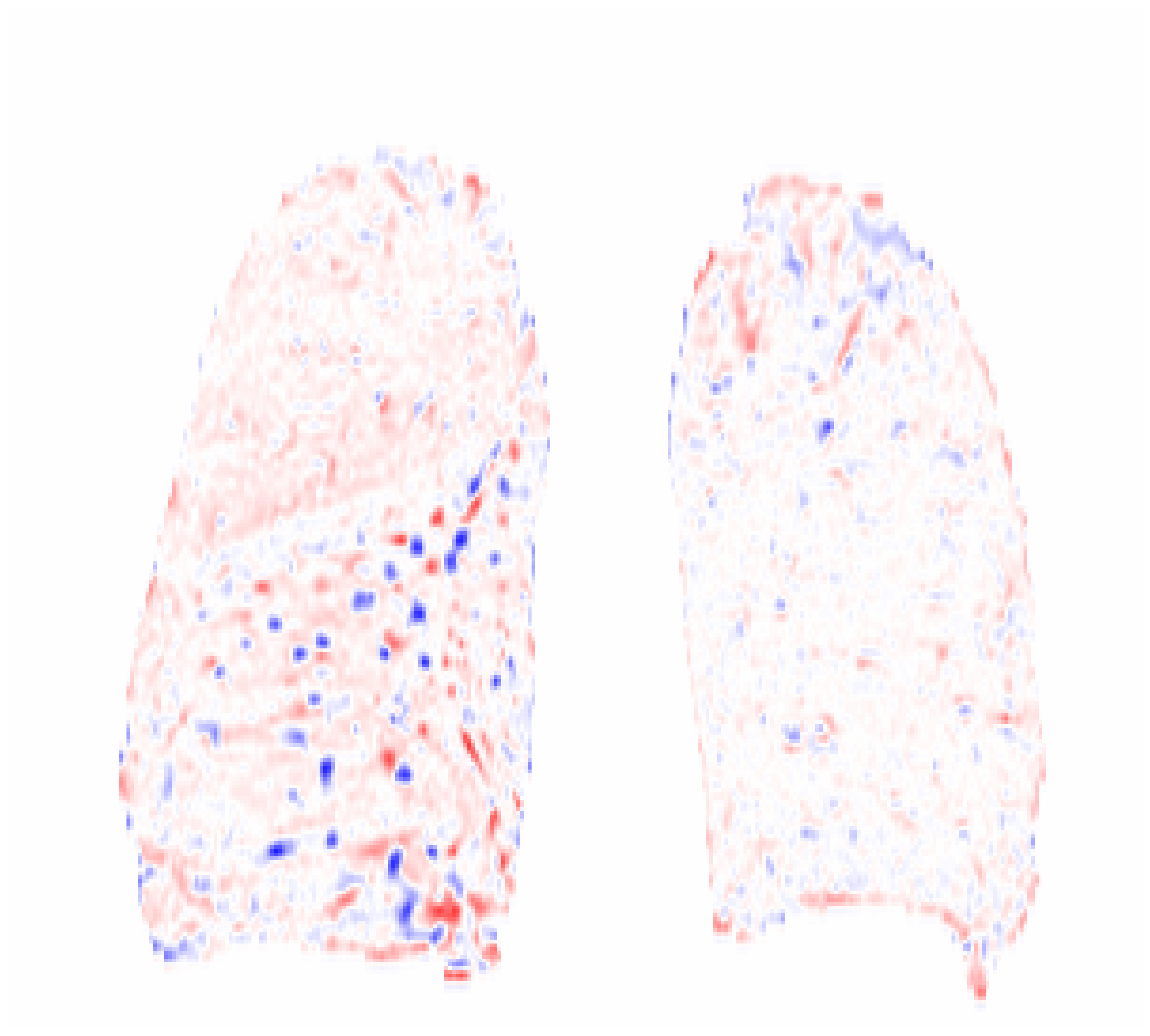}
        \caption{IDIR~\cite{wolterink2022implicit}}
    \end{subfigure}
    \hfill
    \begin{subfigure}[b]{0.19\linewidth}
        \centering
        \includegraphics[width=\linewidth]{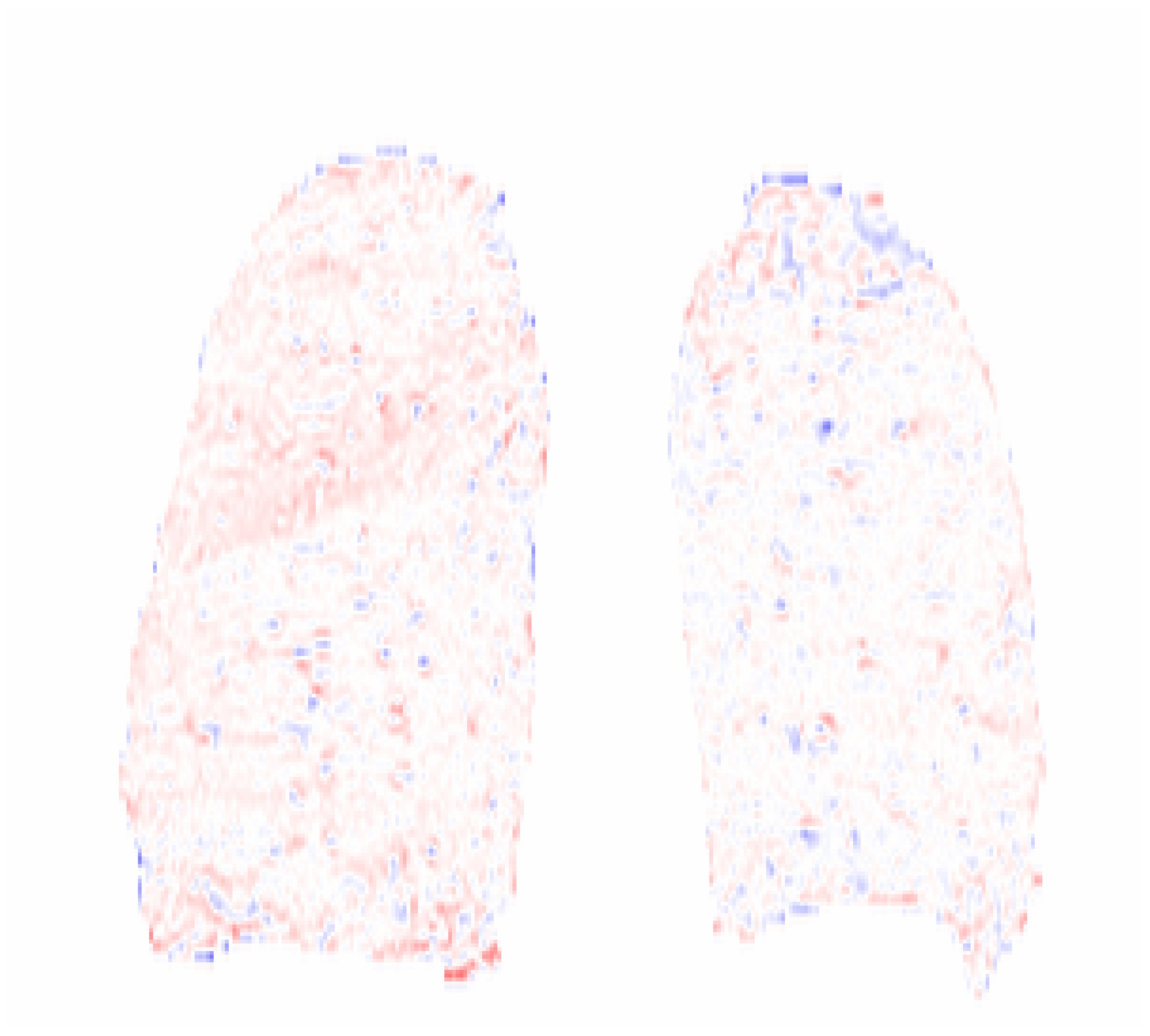}
        \caption{GaussianDIR}
    \end{subfigure}
    \caption{Error maps for Case 8 of DIRLab dataset. This comparison shows error maps generated by four methods, each trained for approximately 2.5 seconds. The depth of the color indicates the magnitude of the error, with deeper colors representing larger errors.}
    \label{fig:DIRLaberrormap}
\end{figure*}

\subsection{Ablation Studies}\label{ablation}
This section evaluates the impact of various design choices in our proposed GaussianDIR, shedding light on their contributions to the overall performance and robustness of the approach. All the ablation experiments are conducted on the OASIS dataset with 49 pairs except the generalization comparison in Sec.~\ref{sec:generalization}.

\paragraph{Free Gaussian} \label{sec:freegaussian}
GaussianDIR leverages fully flexible Gaussian primitives whose positions and shapes can adapt to tissue deformation and structure. We evaluate the impact of using free versus fixed Gaussian primitives on registration accuracy. As shown in Tab.~\ref{tab:ablation}, allowing for free Gaussian primitives yields a $1.4\%$ improvement in DSC and a reduction of $0.06$ in HD95. 

\paragraph{Anisotropic Gaussian and 6-DoF Rigid Transformation}
Isotropic Gaussians, constrained to spherical shapes, are less effective in capturing the nuances of tissue structures. In contrast, anisotropic Gaussians offer a more flexible representation. Tab.~\ref{tab:ablation} shows that employing anisotropic Gaussian primitives improves the DSC from $81.0\%$ to $81.2\%$ and reduces HD95 from $1.92$ to $1.91$. Moreover, introducing a 6-DoF rigid transformation, which combines both rotation and translation, further enhances performance. 

\paragraph{Robustness on Initialization}
We assess the robustness of GaussianDIR to different initialization strategies by comparing random and grid-based initialization. Our experiments reveal that GaussianDIR achieves consistent performance across both methods. A statistical significance test yields a p-value of $0.68$, indicating no statistically significant difference in results between the two strategies at the $5\%$ significance level. 

\begin{table}[htbp]
    \caption{Quantitative evaluation for ablation runs. Aniso., Rot. and Init. denote employing anisotropic Gaussian, 6-DoF rigid transformation rather than translation only, and initialization strategy, respectively.}
    \label{tab:ablation}
    \centering
    \resizebox{1.0\linewidth}{!}{
    \begin{tabular}{c|cc|c|ccc}
    \toprule
        Gaussian & Aniso. & Rot.      & Init.  & DSC (\%) & HD95 & NJD (\%) \\
    \midrule
        Fixed    & \ding{55}  & \ding{55} &  Grid  & 79.6 (2.5) & 1.98 (0.51) & 0.87 (0.23) \\
        Free     & \ding{55}  & \ding{55} &  Grid  & 81.0 (2.3) & 1.92 (0.50) & 1.16 (0.24) \\
        Free     & \ding{51}  & \ding{55} &  Grid  & 81.2 (2.3) & 1.91 (0.51) & 1.22 (0.25) \\
        Free     & \ding{51}  & \ding{51} &  Grid  & 81.3 (2.3) & 1.89 (0.50) & 1.09 (0.24) \\
        Free     & \ding{51}  & \ding{51} & Random & 81.3 (2.3) & 1.89 (0.52) & 1.09 (0.24) \\
    \bottomrule
    \end{tabular}}
\end{table}

\paragraph{Generalization} \label{sec:generalization}
As a case-specific optimization method, GaussianDIR transcends the limitations imposed by training dataset size or domain gaps when adapting to new datasets. To verify its generalization capability, we compared the performance of GaussianDIR and TransMorph~\cite{chen2022transmorph} on the IXI dataset~\cite{ixidataset}. GaussianDIR employes the same set of hyperparameters tuned on the OASIS dataset, while TransMorph uses a model trained on OASIS with weak anatomical map supervision. As shown in Tab.~\ref{tab:generalization}, although Transmorph achieved high accuracy with weak supervision, outperforming GaussianDIR on OASIS with 49 pairs, its performance extremely declined when applied to the unseen IXI dataset, another brain MRI dataset without obvious domain gap as OASIS. 

\begin{table}[htbp]
    \caption{Generalization comparisons between DL-based weak-supervised TransMorph~\cite{chen2022transmorph} and GaussianDIR.}
    \label{tab:generalization}
    \centering
    \resizebox{1.0\linewidth}{!}{
    \begin{tabular}{l|l|ccc}
    \toprule
    Datasets & Methods     & DSC (\%) $\uparrow$ & HD95  & NJD (\%) $\downarrow$ \\
    \midrule
    OASIS    & TransMorph  & 85.7 (1.4) & 1.45 (0.33) & 0.87 (0.21) \\
    OASIS    & GaussianDIR & 81.3 (2.3) & 1.89 (0.50) & 1.09 (0.24) \\
    \midrule
    IXI      & TransMorph  & 69.3 (3.9) & 3.79 (0.69) & 0.91 (0.09) \\
    IXI      & GaussianDIR & 75.6 (2.3) & 2.95 (0.58) & 1.25 (0.16) \\
    \bottomrule
    \end{tabular}}
\end{table}



\begin{figure}[htbp]
    \centering
    \begin{subfigure}[c]{0.48\linewidth}
        \centering
        \includegraphics[width=\linewidth]{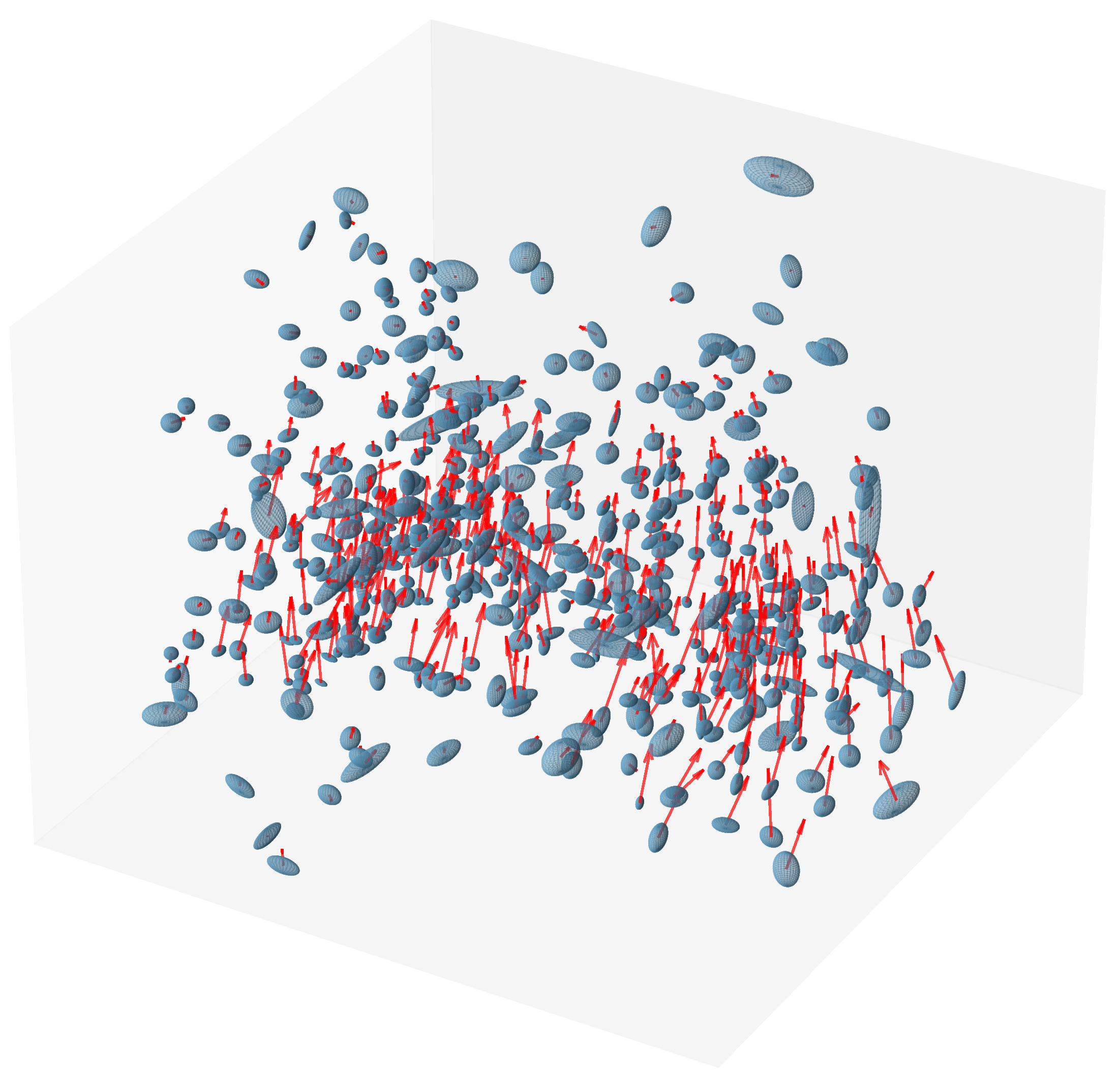}
        \subcaption{}
        \label{fig:gaussian_whole}
    \end{subfigure}
    \hfill
    \begin{subfigure}[c]{0.48\linewidth}
        \centering
        \includegraphics[width=\linewidth]{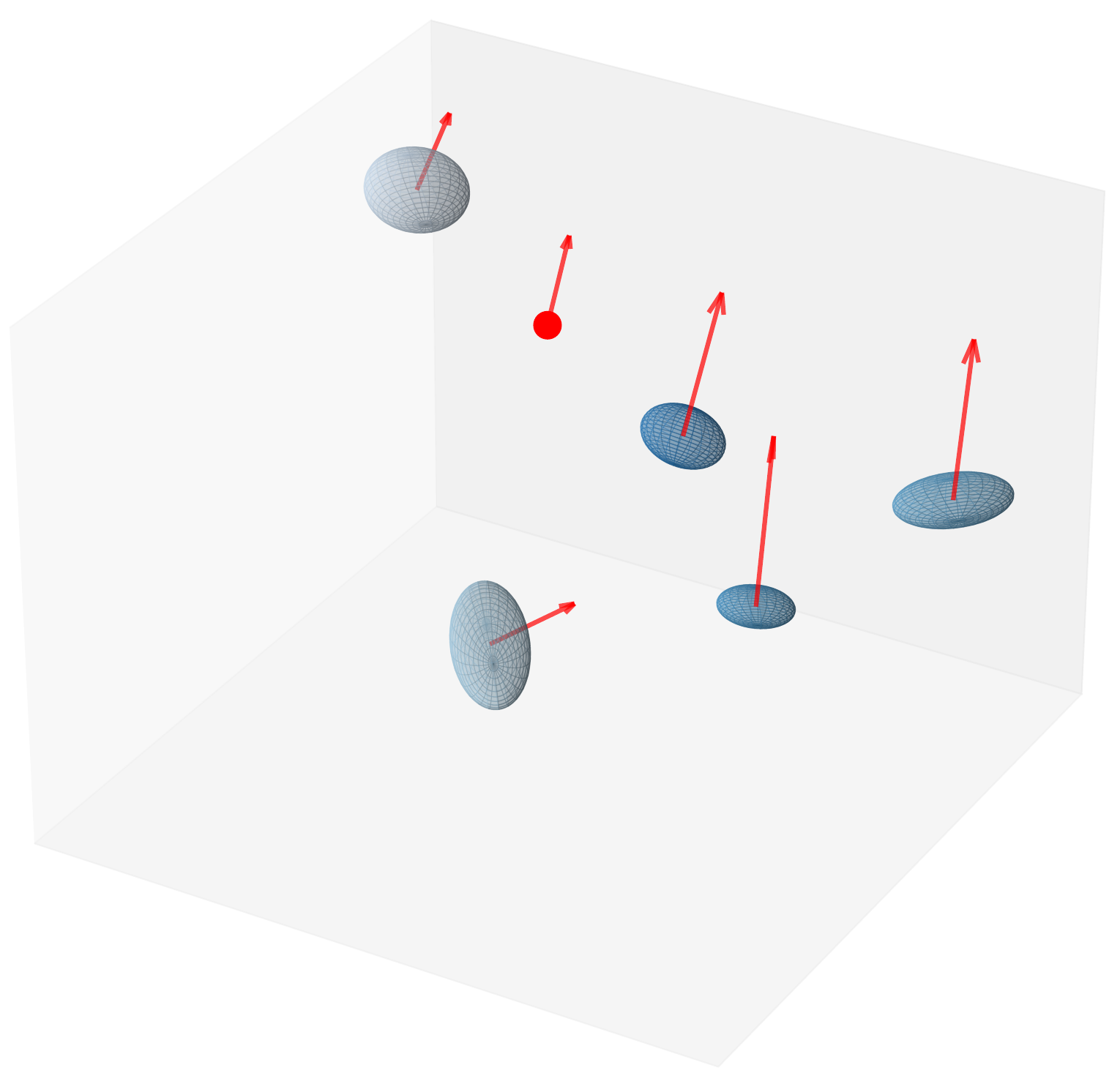}
        \subcaption{}
        \label{fig:gaussian_weight}
    \end{subfigure}
    \caption{Visualization of optimized Gaussian primitives. Blue ellipses and red sphere represent the Gaussian primitives and voxel, respectively. The red arrows denote the displacement vectors of their center positions.}
    \label{fig:gaussian_vis}
\end{figure}

\section{Discussion}

The evaluations in multiple modalities and anatomical regions in our study have demonstrated the superiority of GaussianDIR over classical, DL-based, and INR-based methods. 
Unlike recent advances that predominantly focus on network structures, GaussianDIR offers a novel case-specific optimization solution, achieving higher accuracy within seconds. 
GaussianDIR provides significant benefits, including improved registration accuracy and enhanced computational efficiency, approximating modern demands while challenging the preconception that iterative methods are inherently slow. Our work introduces and validates these advantages across diverse datasets, including brain MRIs, lung CTs, and cardiac MRIs, demonstrating GaussianDIR's robustness and potential for real clinical adoption, such as contour propagation, dose accumulation, respiratory analysis. \jbhi{The experiments on these datasets show the ability of our GaussianDIR to analyze and process mainstream medical images.}

Data-driven and case-specific optimization approaches are two prevailing paradigms in the field of DIR, with the question of which offers the superior solution remaining a topic of ongoing debate~\cite{jena2024deep}.
Data-driven approaches offer real-time inference but require extensive datasets and face challenges in generalization, while case-specific optimization approaches are often thought to suffer from slower convergence. 
As a case-specific optimization method, one of the key advantages of GaussianDIR is its enhanced computational efficiency combined with improved registration accuracy, which is critical for real-time applications in clinical workflows.
For instance, in image-guided surgery and interventional radiology, rapid and precise image registration can dramatically influence patient outcomes by reducing intraoperative delays and enhancing procedural accuracy. 

Beyond efficiency, GaussianDIR demonstrates remarkable robustness and generalization in different modalities and anatomical regions, as evidenced by our results from MRI and CT datasets, including brain, lung, and cardiac images. 
The quantitative comparisons in Tab.~\ref{tab:OASIS-metric-49}, \ref{tab:OASIS-metric}, \ref{tab:DIRLab-metric}, \ref{tab:ACDC-metric} and \ref{tab:generalization} show that GaussianDIR achieves SOTA or comparable performance in these distinct modalities and anatomical structures.
In practical terms, these features mean that clinicians can rely on GaussianDIR in various scenarios, enhancing overall diagnostic accuracy seamlessly. 

Another key advantage of GaussianDIR is its interpretability, distinguishing it from black-box models like DL-based or INR-based approaches. By utilizing explicit Gaussian primitives, GaussianDIR makes the registration process transparent and interpretable for clinicians.
Fig.~\ref{fig:gaussian_whole} demonstrates the Gaussian primitives to aid clinical practitioners in understanding and verifying the deformation fields generated by our model. 
In Fig.~\ref{fig:gaussian_weight}, the ellipses are color-coded to represent blending weights, while the red arrows denote the translation vectors of the Gaussian primitives. These visualizations allow us to evaluate whether the model produces reasonable Gaussian primitives and whether the displacements result from proper blending.
Such transparency is crucial in medical settings, where trust in automated methods depends on clear, verifiable results, thus enhancing confidence in its application for clinical \jbhi{diagnosis and }decision-making.

However, despite these promising results, there remain opportunities for further optimization. One of the current limitations is the \jbhi{automated method, }K-nearest neighbors algorithm, which serves as a bottleneck in the optimization process, impacting overall computational speed. 
Future research could explore organizing the Gaussian primitives using tree structures or implementing tile-based transformation blending with the CUDA programming language, similar 3D GS~\cite{kerbl20233d} and occupancy prediction~\cite{huang2024gaussianformer}, to further enhance efficiency. These enhancements could facilitate even faster registration, further solidifying GaussianDIR’s role in real-time clinical applications.

Another exciting avenue for future research is cross-modality registration. While GaussianDIR has demonstrated excellent performance on both MRI and CT datasets, applying it to cross-modality registration—such as aligning MRI with CT—could unlock valuable insights for treatment planning. 
This integration is crucial for combining complementary information from different imaging modalities, enabling precise anatomical correspondence despite differences in tissue appearance. 
In oncology, CT scans provide detailed electron density maps, while MRI offers superior soft tissue contrast, enhancing tumor visualization and characterization. 

\section{Conclusion}\label{conclusion}
In this paper, we present \jbhi{AI-based} GaussianDIR, a novel training-free automated deformable image registration method that leverages explicit 3D Gaussian primitives to model continuous spatial deformation fields, demonstrating strong interpretability. Our approach simplifies computational demands while ensuring high accuracy and efficiency through the use of free 3D Gaussians and 6-DoF rigid transformations. The adaptive and multi-scale Gaussians effectively manage varying motion complexities and mitigate local minima. 
Empirical validation on the OASIS, DIR-Lab, ACDC, and IXI datasets shows that GaussianDIR outperforms classical and INR-based methods in registration accuracy and efficiency, and it excels in generalization compared to learning-based approaches. This highlights its strong potential for enhancing clinical workflows and improving patient outcomes in diverse medical imaging applications like tumor resection and radiation therapy.

\section*{References}
{\small
  \bibliographystyle{ieee_fullname}
  \bibliography{main}
}


\end{document}